\documentclass{egpubl}
\usepackage{STAG2024}
 
\WsPaper           % uncomment for final version of EG Workshop contribution
\usepackage[T1]{fontenc}
\usepackage{dfadobe}  

\usepackage{cite}  % comment out for biblatex with backend=biber
\BibtexOrBiblatex
\electronicVersion
\PrintedOrElectronic
\ifpdf \usepackage[pdftex]{graphicx} \pdfcompresslevel=9
\else \usepackage[dvips]{graphicx} \fi

\usepackage{egweblnk} 
\usepackage[dvipsnames]{xcolor}
\usepackage{booktabs}
\usepackage{tabularx}
\usepackage{amsthm}
\usepackage{amsmath}
\usepackage{mathtools}
\usepackage{amsfonts}
\usepackage{float}
\usepackage[inkscapeformat=png]{svg}
\usepackage{tikz}
\usepackage{iac_pkg}
\usepackage[linesnumbered,ruled,vlined]{algorithm2e}
\usepackage{graphicx}
\usepackage{subcaption}
\usepackage[labelfont=bf]{caption}
\usepackage{multicol}
\usepackage{adjustbox}
\usepackage[percent]{overpic}
\usepackage{amsmath}
\usepackage[capitalize]{cleveref}
\usepackage{adjustbox}
\usepackage{multirow}
\usepackage{arydshln}
\usepackage{tikz-cd}
\usepackage{lipsum}  
\usepackage[tableposition=top]{caption}
\usepackage{colortbl}

\usetikzlibrary{shapes.geometric, arrows.meta, positioning, decorations.pathreplacing, spy,calc, shadows, shadows.blur}

\tikzstyle{data} = [rectangle, rounded corners, minimum width=2cm, minimum height=1cm, text centered, draw=black, fill=green!30]
\tikzstyle{model} = [rectangle, minimum width=2cm, minimum height=1cm, text centered, draw=black, fill=orange!30]
\tikzstyle{arrow} = [thick,->,>=stealth]

\title[Environment Maps Editing using Inverse Rendering and Adversarial Implicit Functions]{Environment Maps Editing using\\ Inverse Rendering and Adversarial Implicit Functions}

\author[A. D'Orazio, D. Sforza, F. Pellacini, I. Masi]
{\parbox{\textwidth}{\centering Antonio D'Orazio$^1$~\orcid{0009-0009-3046-8812}, Davide Sforza$^1$~\orcid{0000-0002-8042-7156}, Fabio Pellacini$^2$~\orcid{0000-0003-4861-9809}, Iacopo Masi$^1$~\orcid{0000-0003-0444-7646}
        }
        \\
{\parbox{\textwidth}{\centering 
$^1$Sapienza, University of Rome, Italy\\
         $^2$University of Modena and Reggio Emilia, Italy
       }
}
}
\begin{document}

% uncomment for using teaser
\teaser{
      \centering
      \resizebox{0.98\textwidth}{!}{%
        \begin{tabular}{@{}c@{\hskip 0.15em}c@{\hskip 0.15em}c@{\hskip 0.15em}c@{\hskip 0.15em}c@{}}
            \scalebox{1}{Initial Scene} & \scalebox{1}{Target} & \scalebox{1}{Result} & \scalebox{1}{Target} & \scalebox{1}{Result} \\
            % -- ROW 1 --
            \begin{subfigure}{0.2\linewidth} 
                \begin{subfigure}{\linewidth}
                \adjincludegraphics[width=1\linewidth,trim={2.6cm 0.3cm 2.8cm 0.3cm},clip]{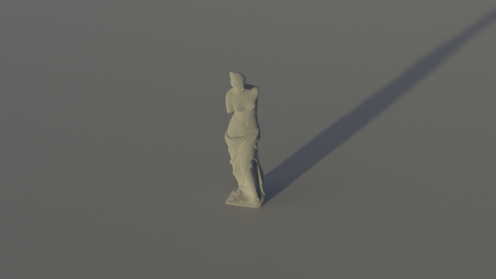}
                \end{subfigure}\par
                \includegraphics[width=\linewidth]{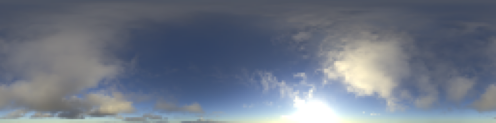}
            \end{subfigure} & 
            % -- ROW 1 --
            \begin{subfigure}{0.2\linewidth} 
                \begin{subfigure}{\linewidth}
                \adjincludegraphics[width=1\linewidth,trim={2.6cm 0.3cm 2.8cm 0.3cm},clip]{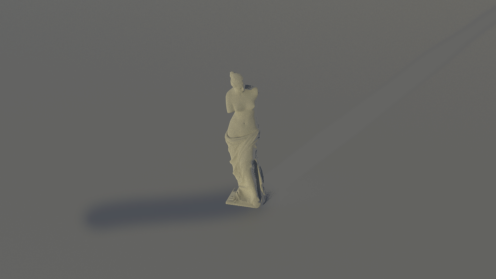}
                \end{subfigure}\par
                \includegraphics[width=\linewidth]{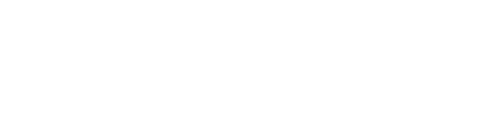}
            \end{subfigure} & 
            % -- ROW 1 --
            \begin{subfigure}{0.2\linewidth} 
                \begin{subfigure}{\linewidth}
                \adjincludegraphics[width=1\linewidth,trim={2.6cm 0.3cm 2.8cm 0.3cm},clip]{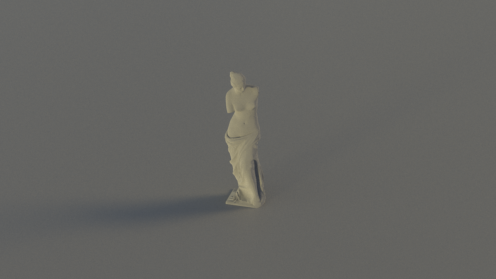}
                \end{subfigure}\par
                \includegraphics[width=\linewidth]{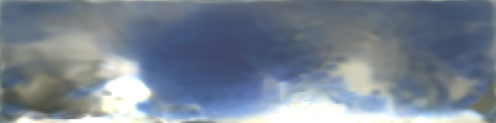}
            \end{subfigure} & 
            % -- ROW 1 --
            \begin{subfigure}{0.2\linewidth} 
                \begin{subfigure}{\linewidth}
                \adjincludegraphics[width=1\linewidth,trim={2.6cm 0.3cm 2.8cm 0.3cm},clip]{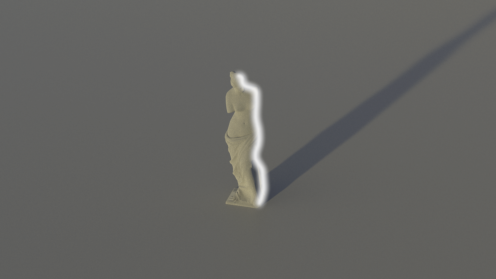}
                \end{subfigure}\par
                \includegraphics[width=\linewidth]{opt_results/placeholder.png}
            \end{subfigure} & 
            % -- ROW 1 --
            \begin{subfigure}{0.2\linewidth} 
                \begin{subfigure}{\linewidth}
                \adjincludegraphics[width=1\linewidth,trim={2.6cm 0.3cm 2.8cm 0.3cm},clip]{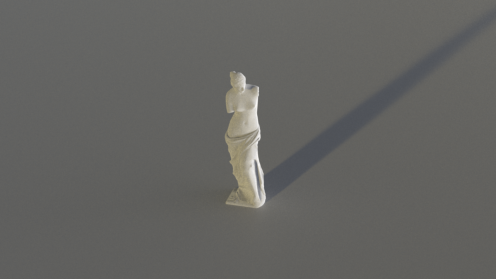} 
                \end{subfigure}\par
                \includegraphics[width=\linewidth]{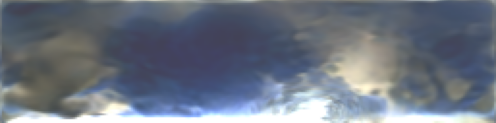} 
            \end{subfigure} \\
         &  & \scalebox{1}{Different Shadow} & &\scalebox{1}{Adding Reflections}  
        \end{tabular}
      }
    
    \caption{We use an adversarial neural environment map combined with an inverse rendering pipeline to generate novel environment maps from a single target rendering. The target can be just edited with a stroke in 2D using image editing software. The figure shows shadow manipulation and adding reflections. The result column displays the rendered results along with the generated environment map.}
    \label{fig:teaser}
    
}

\maketitle
%-------------------------------------------------------------------------
\begin{abstract}
Editing High Dynamic Range (HDR) environment maps using an inverse differentiable rendering architecture is a complex inverse problem due to the sparsity of relevant pixels and the challenges in balancing light sources and background. The pixels illuminating the objects are a small fraction of the total image, leading to noise and convergence issues when the optimization directly involves pixel values. HDR images, with pixel values beyond the typical Standard Dynamic Range (SDR), pose additional challenges. Higher learning rates corrupt the background during optimization, while lower learning rates fail to manipulate light sources. Our work introduces a novel method for editing HDR environment maps using a differentiable rendering, addressing sparsity and variance between values. Instead of introducing strong priors that extract the relevant HDR pixels and separate the light sources, or using tricks such as optimizing the HDR image in the log space, we propose to model the optimized environment map with a new variant of implicit neural representations able to handle HDR images. The neural representation is trained with adversarial perturbations over the weights to ensure smooth changes in the output when it receives gradients from the inverse rendering. In this way, we obtain novel and cheap environment maps without relying on latent spaces of expensive generative models, maintaining the original visual consistency. Experimental results demonstrate the method's effectiveness in reconstructing the desired lighting effects while preserving the fidelity of the map and reflections on objects in the scene. Our approach can pave the way to interesting tasks, such as estimating a new environment map given a rendering with novel light sources, maintaining the initial perceptual features, and enabling brush stroke-based editing of existing environment maps. Our code is publicly available at \href{https://github.com/OmnAI-Lab/R-SIREN}{github.com/OmnAI-Lab/R-SIREN}.  
%-------------------------------------------------------------------------
%  ACM CCS 1998
%  (see https://www.acm.org/publications/computing-classification-system/1998)
% \begin{classification} % according to https://www.acm.org/publications/computing-classification-system/1998
% \CCScat{Computer Graphics}{I.3.3}{Picture/Image Generation}{Line and curve generation}
% \end{classification}
%-------------------------------------------------------------------------
%  ACM CCS 2012
%   (see https://www.acm.org/publications/class-2012)
%The tool at \url{http://dl.acm.org/ccs.cfm} can be used to generate
% CCS codes.
%Example:


\end{abstract}  
%-------------------------------------------------------------------------

\clearpage
\section{Introduction}\label{sec:intro}
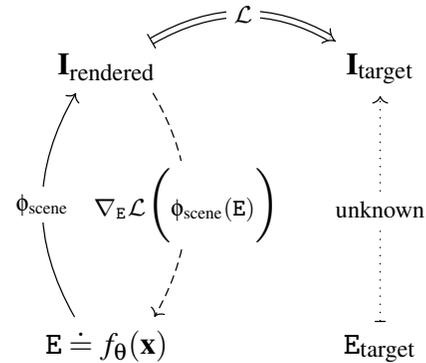
\begin{figure}
\centering
% https://tikzcd.yichuanshen.de/#N4Igdg9gJgpgziAXAbVABwnAlgFyxMJZABgBpiBdUkANwEMAbAVxiRGAB0OAZAQQCUA4gFEABFwC2dHAAsARgDNgASQC+AfU4ccMAB45gAJxhhYxqKsshVpdJlz5CKAEzkqtRizZa+QsZOl5JTVNLh19YBw6QwBzGBxLVWtbEAxsPAIiMgAWd3pmVkR2Ll8RcQ4pWQBjRmBhVXKoCB0AR1EFdTCZeLoACgDZRWBdVQBKJJs7dMciV1zqfK8inwEygZkahjqNLXCDKNj4xOt3GCg4hBRQBUMICSQyEBwIJABGBc9C4o4ACSY48qVDa1bhWagMOhyGAMAAK9gyThAWDA2FgIHByK+TSYcgYrGoUSwDDYUjQcGe6JAUNMSAAzMRJiAbncHgSXohXB4Ct4uDJ-jAuGBIRDQhVApttqp1hLQVw5FgYv0OGgZFhRXtgHAqiYYJYlUCJfVRnKFeNKRCobD4TMisjUfiQAxMWwoHQ4N0oJTqZ7EPTqO6sAocEgALTZRnM+4ctmsrlLb58gFcFVq3Z6AxanXHcGQ6Fw6aZW0orBo6jeumPANB0POCO3KO0mOId6O51FJo4HSegl0IkkuhkikfbnLXn88oaphgADWkAA7mAwY7c1aC4i7SXWKoKKogA

\adjustbox{scale=1.3,center}{
\begin{tikzcd}
{\LARGE \mathbf{I}_{\text{rendered}}} \arrow[rr, "{\Huge \mathcal{L}}" description, Rightarrow, maps to, bend left] \arrow[dddd, "{\huge\nabla_{\envmap}\mathcal{L}\big(\phi_{\text{scene}}(\envmap)\big)}" description, dashed, bend left, shift left=4] &  & {\LARGE \mathbf{I}_{\text{target}}}                                                                      \\
                                                                                                                                                                                                                                                                  &  &                                                                                                          \\
                                                                                                                                                                                                                                                                  &  &                                                                                                          \\
                                                                                                                                                                                                                                                                  &  &                                                                                                          \\
{\LARGE \envmap \doteq f_\theta(\mathbf{x})} \arrow[uuuu, "{\huge \phi_{\text{scene}}}" description, bend left, shift left=2]                                                                                                                                 &  & {\LARGE \envmap_{\text{target}}} \arrow[uuuu, "{\huge \text{unknown}}" description, dotted, maps to]
\end{tikzcd}
}
\caption{We start from an initial environment map $\envmap$ that we parameterize using an implicit function via a robust neural network $f_\theta(\bx)$. 
We optimize the new environment map by receiving gradients through an inverse differentiable rendering that matches the rendered image $\I{rendered}$ to a 
target $\I{target}$. The target is generated by an unknown environment map $\envmap_{\text{target}}$; in practical settings, $\I{target}$ could be created by roughly painting possible shades from $\I{rendered}$ so to ``find'' the generating environment map.
\label{fig:diagram}
\vspace{-5pt}
}
\vspace{-5pt}
\end{figure}

\begin{figure*}[h]
\begin{center}
\begin{adjustbox}{max width=\textwidth}
\begin{tikzpicture}[node distance=2.5cm, auto]

% Define the styles for the nodes and arrows
\tikzstyle{data} = [rectangle, rounded corners, minimum width=2.5cm, minimum height=1.5cm, text centered, draw=gray, fill=cyan!15, blur shadow={shadow blur steps=5}]
\tikzstyle{process} = [rectangle, rounded corners, minimum width=3cm, minimum height=1.5cm, text centered, draw=gray, fill=orange!15, blur shadow={shadow blur steps=5}]
\tikzstyle{model} = [rectangle, rounded corners, minimum width=3cm, minimum height=1.5cm, text centered, draw=gray, fill=green!15, blur shadow={shadow blur steps=5}]
\tikzstyle{arrow} = [thick,->,>=stealth, color=gray]
% Nodes with shadow effects
\node (coords) [data] {\parbox{2.5cm}{\centering{\LARGE $\bx \doteq (x, y)$}\\ $\I{hdr}(\bx)$ coordinates}};
\node (siren) [model, right of=coords, node distance=4cm] {\parbox{2.5cm}{\centering {\LARGE$$f_{\net}(\bx)$$} SIREN \\ Neural Network}};
\node (denorm)[process, right of=siren, node distance=4cm] {{\large De-Normalize w/ \cref{eps_norm}}};
\node (expo)[process, right of=denorm, node distance=4cm]{{\large Exponential w/ \cref{eq:eps_normexp}}};
\node (hdr) [data, right of=expo, node distance=4cm] {
    \parbox{2.5cm}{\centering
    \includegraphics[width=2.5cm]{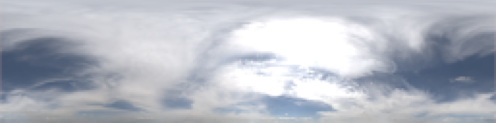} \\ HDR Env. Map}};
\node (render) [process, right of=hdr, node distance=4cm] {\parbox{2.5cm}{\centering {\LARGE$$\phi_{\text{scene}}$$} Render scene}};
\node (rendering) [data, right of=render, node distance=4cm] {
    \parbox{2.5cm}{\centering
    \includegraphics[width=2.5cm]{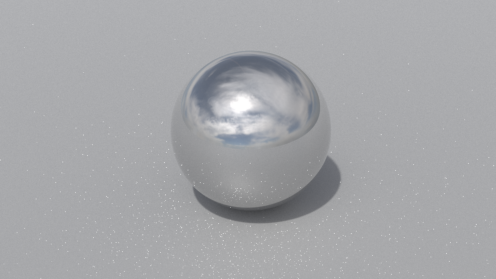} \\ 
    Current Rendering}};
\node (loss) [process, below of=rendering, node distance=3cm] {\parbox{2.5cm}{\centering {\LARGE$$\Loss$$} Loss}};
\node (target) [data, below of=render, node distance=3cm] {
    \parbox{2.5cm}{\centering
    \includegraphics[width=2.5cm]{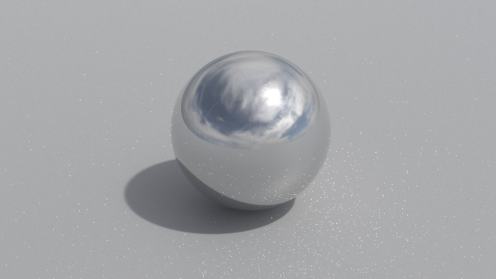} \\ 
    Target Rendering}};

% Arrows
\draw [arrow] (coords) -- (siren);
\draw [arrow] (siren) -- (denorm);
\draw [arrow] (denorm) -- (expo);
\draw [arrow] (expo) -- (hdr);
\draw [arrow] (hdr) -- (render);
\draw [arrow] (render) -- (rendering);
\draw [arrow] (rendering) -- (loss);
\draw [arrow] (target) -- (loss);
\draw [arrow] (loss) -| ++(2,1) |- ++(0,3.5) -| node[midway, above] {{\Large Update Weights}} (siren);

% Brace with text
\draw [decorate, decoration={brace, mirror, amplitude=10pt, raise=5pt}, thick] 
([yshift=-0.1cm]siren.south west) -- ([yshift=-0.3cm]expo.south east) 
node[midway, yshift=-36pt]{\LARGE SIREN HDR};

\end{tikzpicture}
\end{adjustbox}
\end{center}
\caption{Inverse rendering in the proposed SIREN HDR}
\label{fig:pipeline}
\end{figure*}

High-quality environment maps are essential for achieving photorealistic renderings.
However, they are often treated as static assets, limiting the ability of artists to customize the environment. Instead of editing existing maps, artists typically choose from a pre-curated collection.
In light of this, our work aims to present an approach to perform the editing of environment maps by implementing a differentiable rendering in the pipeline, allowing backpropagation from the final rendered image to the initial objects in the 3D scene. This approach can be seen as an inverse problem, which can be solved via optimization techniques such as gradient descent to iteratively update the color values of the environment maps.
Instead of optimizing the map directly in the pixel space, which is highly fragile, or resorting to expensive generative models based on diffusion process~\cite{lyu2023diffusion}, we take a lazy learning approach and parameterize each environment map using a robust implicit function implemented using a standard multi-layer perception (MLP). To attain naturally looking images, we train our robust function so that for small perturbations of the parameters---the gradients received by the inverse rendering---the output of the new map remains smooth and stable.
\cref{fig:diagram} shows an overview of our approach: a target image $\I{target}$ is rendered from an \emph{unknown} environment map. We fit an implicit function $f_\theta(\mathbf{x})$ to an initial environment map $\envmap$. The method renders the $\envmap$ and updates its representation to match the target image. Instead of using plenty of regularizers to constrain the problem, we make our implicit function smooth and robust so that once we update its weights, the new map looks natural and has fewer artifacts. When designing our implicit representation, we employed SIREN \cite{sitzmann2019siren}, which has been proven to reconstruct images with very high fidelity, and adapted its training strategies to correctly represent HDR images, which was not possible in the original approach. We call this method HDR SIREN.~\cref{fig:pipeline} shows how it is integrated into the inversion algorithm.
Our method is designed to allow computer graphics artists to fine-tune the lighting of 3D scenes by directly manipulating the environment maps, which traditionally has been a complex and time-consuming process. The key objective is to enable precise adjustments to light sources within these maps without corrupting the background, maintaining fidelity in reflections and lighting. Our method can be applied to downstream tasks, as shown in ~\cref{fig:teaser}, in which we alter the appearance of the environment map by specifying coarse sketches over the initial rendering.
% %%%%%%%%%%%%%%%%%%%%%%%%%%%%%%%%%%%%%%%%%%%%%%%%%%%%%%%%%%%%%
% %%%%%%%%%%%%%%%%%%%%%%%%%%%%%%%%%%%%%%%%%%%%%%%%%%%%%%%%%%%%%

During our work, we faced the following challenges: $\diamond$ \emph{Sparsity of Relevant Pixels}: The optimization of environment maps is hampered by the sparsity of relevant pixels. In most cases, only a small fraction of the image directly contributes to illuminating the scene, making it difficult to achieve consistent and meaningful updates. This sparsity leads to noise and convergence issues, especially when dealing with HDR images; $\diamond$ \emph{Convergence Difficulties with HDR Images:} High Dynamic Range (HDR) images present significant challenges during optimization. The wide range of pixel intensities in HDR images, which can span from zero to infinity, complicates the optimization process. High learning rates may cause artifacts in the background, while low learning rates fail to effectively manipulate the light sources, leading to suboptimal results.; $\diamond$ \emph{Balancing Background and Light Source Manipulation}: A key challenge lies in balancing the optimization between maintaining the integrity of the background and manipulating the light sources. Techniques such as separating the light sources from the background and applying regularization methods are necessary to prevent the background from being corrupted while still allowing effective light source modification; $\diamond$ \emph{Noise Reduction}: The optimization process often introduces noise into the environment maps, particularly when using plain optimization with no regularization. This noise must be mitigated through regularization techniques like Total Variation regularization, which penalizes rapid changes in pixel values, ensuring a smoother and more coherent output; $\diamond$ \emph{Maintaining Luminance Consistency}: During the optimization process, it is crucial to maintain the overall luminance of the environment map to preserve the correct balance between light sources and shadows. This requires careful management of luminance regularization to ensure that the optimized environment map does not deviate significantly from the original in terms of brightness and contrast.

Given the challenges mentioned above, our work aims to provide the following contributions: $\diamond$ \emph{Inverse rendering-based HDR optimization pipeline}: We propose a lazy-learning optimization pipeline that reconstructs a desired environment map given a target rendering. In particular, we explore how to constrain the sparsity and the high variance in values of HDR images to correctly converge to a new HDR image, maintaining the overall consistency of the sky environment. Such exploration involves regularization techniques, e.g. defining a penalty on the total luminance, and image representations that help the Adam optimizer to reach convergence. We will focus on two image representations: 1) pixel-based representations, such as optimizing the environment map in the logarithmic space; 2) implicit representations that map the $(x,y)$ image grid to colors using a neural network, representing HDR images. Our representation is robust to gradient updates received from the inverse rendering, thereby providing smooth editing of environment maps;
$\diamond$ \emph{Improvements over the SIREN implicit representation}: The baseline for the SIREN implicit representation does not converge with HDR images. In this work, we show how to correctly train a neural network based on periodical activation functions such that it can encode high spikes in the image. Finally, we show how to enforce robustness in the weights space to ensure that the result of the inverse rendering-based optimization pipeline will produce natural-looking images.

The next section reviews the related research. \cref{sec:img} explains the image representations used in our work, divided between traditional RGB pixel representation and implicit representations. \cref{sec:R-SIREN} describes our proposed robust implicit representation and the process to fit it to an HDR map, while \cref{sec:rendering} describes the inverse rendering process and the optimization pipeline and how it tackles the challenges addressed earlier. Finally, \cref{sec:expt} covers the training and implementation details, presents the results, and includes a discussion of those results.
%-------------------------------------------------------------------------
\section{Related Work}\label{sec:related}
\minisection{Appearance Editing} \emph{EnvyLight}~\cite{Pellacini2010envyLightAI} uses an algorithm to select environment map regions by sketching strokes on lighting features in the rendered image, separating the background and the foreground features in the environment map. The paper proposes the use of wavelets to capture all frequency effects. The separation between background and foreground is proposed again in this work. Instead of being specified manually by the user, the separation is performed automatically during the optimization process. \emph{Illumination Brush: Interactive Design of All-frequency Lighting} \cite{4392727} is another traditional approach for environment maps editing for which the authors designed a 3D viewport where the user can paint over the surface of a 3D model desired lighting. Then, it estimates the environment map, creating the lighting style as an inverse problem. The final environment map, however, does not resemble a captured photograph. \emph{Appearance editing with free-viewpoint neural rendering} \cite{gera2021appearance} proposes a method to estimate a BRDF using a differentiable renderer and a few reference images at different angles, enabling material and shading edits on 3D models, showing the potential of differentiable rendering for appearance editing.

\minisection{Inverse Rendering and Model Inversion} Differentiable rendering is a broad field, where the work can involve different parts of the pipeline. Mainly, the work can be grouped as solving optimization problems and working on the differentiable renderer itself. \emph{Large Steps in Inverse Rendering}~\cite{Nicolet2021Large} is a breakthrough method that dramatically improves the reconstruction of 3D meshes from a target image, giving many perspectives on how to tune the optimization function. The idea of using the Laplacian regularizer, which is translated to the Total Variation (TV) regularizer for simpler domains like 2D images, helped with noise and convergence.
Developed by the same research group, \emph{Mitsuba 3} \cite{Jakob2020DrJit} is the differentiable rendered that they used to implement their work, built on top of the just-in-time compiler called Dr.Jit. It is a retargetable renderer, meaning that the data structures can be adapted to perform different tasks. For example, the autograd mode allows the renderer to keep track of the gradients with respect to the input parameters, such as textures, geometries, and material parameters. Given that the environment map is an image texture, the native support for keeping gradients for textures was the reason for choosing Mitsuba over other differentiable renderers. Another example of a differentiable renderer is represented by the \emph{NVIDIA Kaolin}~\cite{KaolinLibrary} suite, which is a PyTorch library for simplifying the work with 3D assets. However, their differentiable renderer is more suitable for optimizing 3D shapes and light positions. Therefore, Mitsuba 3 resulted in a better choice for our work. Finally, the work in~\cite{wang2021imagine,rouhsedaghat2023magic} are one-shot methods for image synthesis that use structured gradients received from the inversion of a pre-trained classifier to control manipulations; ~\cite{rouhsedaghat2023magic} moreover employs a robust model to receive perceptually aligned gradients (PAG) and allows the user to edit the image using a binary mask. Finally, \cite{lyu2023dpi} proposes a similar pipeline compared to our approach, but, instead, they involve diffusion posterior sampling as a prior for the inverse rendering process, achieving remarkable results but requiring an expensive pre-training of a Denoising Diffusion Probabilistic Model (DDPM)~\cite{ho2020denoising} over multiple environment map samples. Finally, \textit{Photorealistic rendering of mixed reality scenes} \cite{Kronander2015PhotorealisticRO} provide an extensive survey of differentiable rendering approaches involving HDR illumination.

\minisection{Adversarial Robustness} Neural network robustness has increasingly become a central topic in deep learning. The work shown in AT \cite{madry2017towards}, which incorporates adversarial examples into the training process, is still the most effective empirical strategy, with many modifications developed on top of it. Zhang \cite{zhang2019theoretically} proposed the TRADES loss, which leverages the Kullback-Leibler (KL) divergence to balance the trade-off between standard and robust accuracy. Cui \cite{cui2023decoupled} improved the KL divergence, addressing its asymmetry property with improved results. Finally, Adversarial Weight Perturbation (AWP) \cite{Wu2020AdversarialWP} shows how to perturb the training weights, helping the generalization by flattening the loss landscape. Unlike classic adversarial perturbation in the input space, AWP is scaled proportionally to the relative size of the original weights for each layer. Finally, robust classifiers are mainly known for their discriminative power, nevertheless prior art~\cite{mirza2024shedding} also linked them to generative models.

%-------------------------------------------------------------------------
\section{Image Representations}\label{sec:img}

\minisection{HDR RGB} HDR imaging is used to accurately represent scenes with a wide range of light intensities. Unlike traditional 8-bit RGB, HDR RGB encodes luminance using higher bit depths (e.g., 16-bit or 32-bit), capturing more detail in both bright and dark areas. This representation is crucial for the environment maps, as it preserves the full spectrum of light in a scene, allowing the representation of the full intensity of the light sources. HDR images are theoretically coded in the range: $$\I{hdr}(\bx) \in [0, +\infty) \qquad \bx = (x,y) \in \mathbb{R}^2$$
Where the practical upper limit is determined by the specific HDR implementation. SDR images are represented in the range: $$\I{sdr}(\bx) \in [0, 1], \quad \bx = (x,y) \in \mathbb{R}^2$$
This image representation is the most trivial one, but it requires additional image processing and regularization to constrain the inverse optimization problem. Also, the large range in values of HDR images is not well-suited for gradient descent optimization. Higher learning rates will optimize the few light sources correctly at the cost of ruining the remaining background. Lower learning rates will optimize only the background pixels, keeping the light sources substantially unchanged.

\minisection{SIREN Implicit Representation}
Optimizing directly on the environment map pixel will lead to low-quality images due to the sparsity nature of the problem, producing new images that are off the manifold of the natural images. To represent the environment map, we first train a SIREN (Sinusoidal Representation Networks) \cite{sitzmann2019siren} model to implicitly represent the image. The SIREN model leverages periodic activation functions to capture the high-frequency details in the image. Specifically, the image is represented as a continuous function $f_\theta: \mathbb{R}^2 \rightarrow \mathbb{R}^3$, where $\theta$ denotes the network parameters. Given a spatial coordinate $\mathbf{x} = (x, y)$, the network outputs the corresponding RGB color values $f_\theta(\mathbf{x}) = (r, g, b)$. The sine activation function is defined as $\sigma(\mathbf{x}) = \sin(\omega_0 \mathbf{x})$, 
where $\omega_0$ is a frequency parameter controlling the period of the sine function. The first layer processes the input spatial coordinates as follows $\mathbf{h}_1 = \sin(\omega_0 \mathbf{W}_0 \mathbf{x} + \mathbf{b}_0)$. Then, each layer $i$-th computes:
$$\mathbf{h}_{i+1} = \sin(\mathbf{W}_i \mathbf{h}_i + \mathbf{b}_i)$$
where $\mathbf{W}_i$ and $\mathbf{b}_i$ are the weight matrix and bias vector of the $i$-th fully connected layer. All the sets of parameters are thus $\net \doteq \{\mbf{W}_i,\mbf{b}_i\}$ and the network is a multi-layer perceptron (MLP). This configuration allows the SIREN model to map spatial coordinates directly to pixel values, although additional care is needed to correctly model HDR images.

\section{Representing HDR Images with SIREN}\label{sec:R-SIREN}
\begin{figure}[b]
\centering
\captionsetup[subfigure]{width=0.8\textwidth}
\captionsetup[figure]{width=0.8\linewidth}
\centering
\begin{subfigure}{.98\columnwidth}
    \begin{overpic}[scale=0.6]{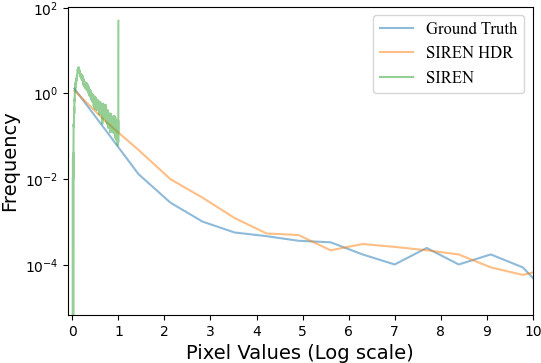} 
        \put(0,0){\color{white}\rule{82.1mm}{3.1mm}} 
        \put(0,0){\color{white}\rule{3.12mm}{50mm}} 
        \put(35.5,0){\color{black}{\small Pixel Values (Log scale)}} 
        \put(0,26.5){\rotatebox{90}{\color{black}{\small Frequency}}}  
    \end{overpic}
\end{subfigure}~\par\smallskip
\vspace{0.4em}
\begin{subfigure}{.33\columnwidth}
  \includegraphics[width=\linewidth]{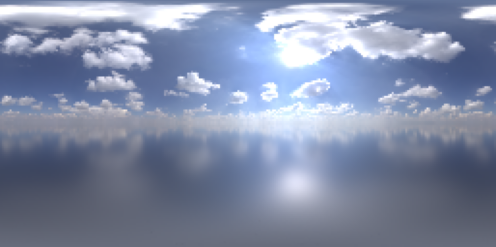}
    \caption*{Ground Truth}
\end{subfigure}~%
\begin{subfigure}{.33\columnwidth}
  \includegraphics[width=\linewidth]{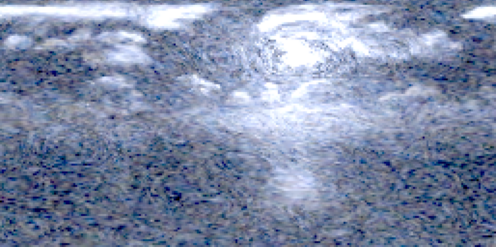}
    \caption*{SIREN}
\end{subfigure}~%
\begin{subfigure}{.33\columnwidth}
  \includegraphics[width=\linewidth]{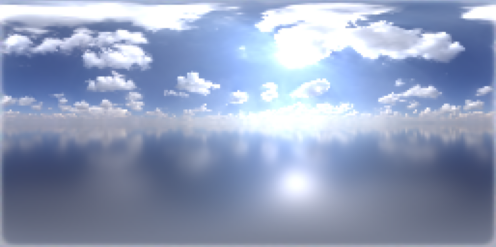}
    \caption*{SIREN HDR}
\end{subfigure}%
\caption{SIREN HDR can represent HDR images with higher fidelity and better pixel intensity distribution, compared to the baseline architecture in \cite{sitzmann2019siren}.}
\label{fig:histogram}
\end{figure}
The standard definition of SIREN does not model HDR images directly, as the sinusoidal activation functions can reconstruct only the standard color range before losing its learning capabilities. To address this, we train SIREN to represent the environment map in the logarithmic space. Given a target HDR environment map $\envmap$, we train SIREN to represent the normalization in \cite{Xu2005HighdynamicrangeSE}:
\begin{equation}\label{eq:siren_pred}
f_\theta(\bx) \approx \frac{\log(\envmap) - \min(\log(\envmap))}{\max(\log(\envmap))-\min(\log(\envmap))}.
\end{equation}
Then, to obtain the learned environment map $\envmap_{\text{pred}}$, we de-normalize and transform it back to the HDR space as:
\begin{align}\label{eps_norm} 
    \widehat{f}_\theta(\bx) &=   f_\theta(\bx) \big(\max(\log(\envmap)) - \min(\log(\envmap))\big) + \min(\log(\envmap)) 
\end{align}
\begin{align}\label{eq:eps_normexp}
    \envmap_{\text{pred}} &=  \exp\big({\widehat{f}_\theta(x)}\big) 
\end{align}
Note that, to prevent numerical instability, we add a small value $\epsilon = 0.01$ to the pixels inside the logarithm, which is omitted for readability.

To capture the few high peaks of an HDR image, and thus address the problem of sparsity, we enforce this behavior in the loss function. We define two loss functions as:
\begin{multline}\label{eq:loss}
\hspace{-1em}
\Loss = \underbrace{\lambda_{\text{log\_mse}}\norm{f_\theta(\bx) - \log(\envmap)} + \lambda_{\text{log\_ssim}}\text{SSIM}\big(f_\theta(\bx),~\log(\envmap)\big)}_{\Loss_{\text{log}}}+\\+ \underbrace{\lambda_{hdr}\norm{\hat{f}_\theta(\bx) -  \envmap}} _{\Loss_{\text{hdr}}}
\end{multline}
The loss $\Loss_{\text{log}}$ in the log space will focus more on reconstructing the overall features of the environment map. In addition, by defining a loss function in the re-transposed HDR space, e.g. $\Loss_{\text{hdr}}$, the optimizer can give more importance to the few pixels with high peaks, therefore addressing the sparsity issue.~\cref{fig:histogram} shows how such strategies result in better coverage of the whole HDR space.

\minisection{Enabling Smooth Editing via Adversarial Perturbation} We propose to interpret the weights $\theta$ of neural implicit representations as a form of ``embedding'' for the target image. With this interpretation in mind, each specific set of weights corresponds to a unique representation of the data. This raises the question of whether we can adjust these weights such that perturbations do not disrupt the underlying image space. If we do so we could navigate ``around the embedding'' to synthesize new natural-looking images. Conceptually, this can be viewed as exploring whether the weight set belongs to a smooth implicit latent space, which is not directly observable. Adversarial training \emph{in the weight space}~\cite{Wu2020AdversarialWP} enables us to craft this smooth implicit latent space. We thus seek for a set of weights $f_\theta(x)$ such that for small perturbation in the weights space, the implicit representation remains stable and outputs natural environment maps. Once we have fit the MLP and run the optimization, our implicit representation is robust to gradient updates we receive from the inverse rendering.

\begin{figure*}[tb]
  \centering
  \resizebox{\linewidth}{!}{%
     \begin{tabular}{@{}c@{\hspace{0.045in}}c@{\hspace{0.045in}}c@{\hspace{0.045in}}c@{\hspace{0.045in}}c@{}}
       Original & SIREN HDR & R-SIREN HDR & SIREN HDR~+~$\mbf{W}^{\star}$ & R-SIREN~HDR~+~$\mbf{W}^{\star}$ \\  
      \includegraphics[width=0.2\linewidth]{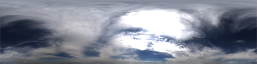} &
      \includegraphics[width=0.2\linewidth]{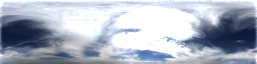} &
      \includegraphics[width=0.2\linewidth]{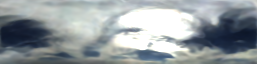} &
      \includegraphics[width=0.2\linewidth]{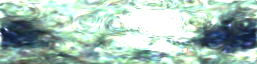} &
      \includegraphics[width=0.2\linewidth]{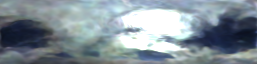} \\
      \end{tabular}
  }
\caption{Effects of applying the \emph{same} small random perturbation to implicit functions. The random perturbation is $\mbf{W}^{\star}\sim \alpha\mathcal{N}(0,1)$ with $\alpha$ set to \texttt{$1\times 10^{-3}$} and is applied in the weights space. From left to right: the original environment map, the rendering of SIREN HDR, the rendering of R-SIREN HDR, the perturbation applied to SIREN HDR, and, finally, the perturbation applied to R-SIREN HDR. In this last case, perturbation of the weights induces naturally looking maps.}
\label{fig:perturbation}
\end{figure*}

In practical terms, our approach involves enforcing robustness by incorporating adversarial perturbations during the training process. Specifically, if we consider a specific layer \( \mbf{W} \), we seek to optimize it to minimize the worst-case loss across potential perturbations \( \Delta\mbf{W} \in \Omega \). This is formalized as the following optimization problem:
\begin{equation}
\min_\mbf{W} \max_{\Delta\mbf{W} \in \Omega} \rho(\mbf{W} + \Delta\mbf{W}) \rightarrow \min_\mbf{W} \max_{\Delta\mbf{W} \in \Omega} \frac{1}{n} \sum_{i=1}^{n} \Loss(f_{\mbf{W} + \Delta\mbf{W}}(\mbf{x}_i, \I{hdr}(\bx)_i))
\end{equation}
In this expression, \( \rho(\mbf{W} + \Delta\mbf{W}) \) represents the robust loss function, \( f_{\mbf{W} + \Delta\mbf{W}} \) denotes the model with perturbed weights, \( \Loss \) is defined in \cref{eq:loss}, \( \mathbf{x}_i \) and \( \I{hdr}(\bx)_i \) are the input grid and the corresponding target image. Unlike the classic adversarial examples in the input space where each input is perturbed in a fixed neighborhood, in our case, the feasible set of the perturbation $\Omega$ is restricted using each relative weight size. Thus $\Omega = \{ \forall~\Delta : \norm{\Delta} \leq \gamma \norm{\mbf{W}} \}$ and $\gamma$ is a fixed constant used to bound the weight perturbation. Note that, unlike~\cite{Wu2020AdversarialWP}, we \emph{just} perturb the weights space and not the input space. ~\cref{alg:awp} illustrates the practical approach to perform adversarial weight perturbation, while its role during the training can be seen in ~\cref{alg:training}. We will refer to SIREN neural networks trained with the explained robustness strategy as R-SIREN. The effects of training implicit representations with robustness can be seen in ~\cref{fig:perturbation}.
    
\begin{algorithm}
\SetAlgoLined
\KwIn{Set of weights $\theta$, input coordinates $\bx$}
\KwOut{Perturbation $\theta^*$}

$\hat{\theta} \leftarrow \text{clone } \theta$\\

$Y \leftarrow f_{\hat{\theta}}(\bx)$\;
$\Loss_{\text{awp}} \leftarrow -\Loss_{\text{SSIM}}(Y, \envmap)$\

$\text{update weights } \hat{\theta}$\\
$\Delta\theta \leftarrow \theta-\hat{\theta}$\\

$\theta^*\leftarrow \frac{\|\theta\|}{\|\Delta\theta\| + \epsilon} \cdot \Delta\theta$\\

\caption{Computing the Adversarial Weight Perturbation (calc\_awp)}
\label{alg:awp}
\end{algorithm}
\begin{algorithm}
\SetAlgoLined
\KwIn{Target environment map $\envmap$, model $f_\theta$, input coordinates $\bx$, learning rate $\eta$, number of iterations $N$}
\KwOut{Trained model $f_\theta$}

\For{$i = 1$ \textbf{to} $N$}{
    $\theta^* \leftarrow \text{calc\_awp}(\theta_{i-1})$\\
    $\theta_{adv} \leftarrow \theta_{i-1}+\theta^*$\\

    $Y \leftarrow f_\theta(\bx)$\;
    $\Loss_{\text{robust}} \leftarrow  \Loss_{\text{log}}(Y, \envmap) + \Loss_{\text{hdr}}(Y, \envmap)$\

    $\text{update weights } \theta_{adv}$\\
    $\theta_i \leftarrow \theta_{adv}-\theta^*$\\

}

\caption{Training SIREN with Adversarial Weights Perturbation}
\label{alg:training}
\end{algorithm}

\section{Inverse Rendering Optimization}\label{sec:rendering}
Given a target image $\I{target}$, we want to minimize the loss function between the rendered image $\phi_{\text{scene}}(\envmap)$ and $\I{target}$, by iteratively altering the environment map $\envmap$ as: 
\begin{multline}\label{fig:main_opt}
    \min_{\envmap}  \Loss(\I{rendered}, ~\I{target}) + \Loss_{\text{reg}}(\envmap) = \\ = \Loss(\phi_{\text{scene}}(\envmap), ~\I{target}) + \Loss_{\text{reg}}(\envmap)
\end{multline}
The ill-posed nature of inverse rendering problems often leads to trivial solutions that minimize error without preserving meaningful image features. Unconstrained optimization can exploit these ambiguities, for instance, by manipulating only reflective regions or darkening the brightest pixels. This can result in a loss of smooth transitions, a property found in natural images. To address this, we employ regularization techniques to constrain intensity, luminance, and perceptual characteristics, ensuring that the synthesized image is consistent with the initial sample.

\subsection{Regularization}

\minisection{Luminance Regularization} During the optimization process, we want to keep the overall luminance of the environment map constant to maintain the correct balance between light sources and shadows. To achieve this result, we propose a luminance regularization that aims to balance the difference in luminance between the original environment map and the optimized one. We can compute the luminance as:
\begin{equation}\label{lum}
\text{Lum}(\envmap) = \sum_{i,j} \sum_{k \in \{\text{r,g,b}\}} \alpha_k \envmap_{i,j}[k].
\end{equation}
where $\alpha_r = 0.2126, \alpha_g = 0.7152, \alpha_b = 0.0722$, per definition of luminance and $\envmap_{i,j}[k]$ represents the $k$-th color plane of the  environment map at pixel $i, j$.
Subsequently, the proposed luminance regularization is:
\begin{equation}\label{eq:reg_lum}
    \Loss_{lum} = \lambda_{lum} | \text{Lum}(\envmap) - \text{Lum}(\envmap^{\prime}) |
\end{equation}
Where $\envmap^{\prime}$ is the initial environment map, and $\lambda_{lum}$ is the trade-off parameter.
During the rendering process, usually, only the upper part of the image is used because the lower part is covered by the scene elements. When using regularization techniques such as luminance regularization, some scene configurations, such as the presence of the infinite plane, could cause the optimizer to cheat by simply lowering the brightness of the lower part of the environment map, keeping the amount of luminance constant but burning the usable section. To avoid this from happening, the optimization is run only on the higher part.

\minisection{Extracting Lights Information}
Once the global luminance is constrained, the optimizer can cheat by lighting up a large amount of pixels to avoid penalties from the luminance regularization. However, environment maps usually have a small set of pixels that are very large in value, surrounded by values in the standard range. For example, in the presence of the sun. Finding the pixels belonging to a light source in an entire environment map is a \emph{sparse} problem. For instance, in a traditional 1K environment map, we have a total of $1024 
\times 512 = 524,288$ pixels. On the other hand, the light sources, like the sun, usually represent only a few dozen of the whole image. 
Therefore, we want to encourage sparsity by applying the $\ell_1$ regularization over the environment map values. Such regularization, when competing against $\Loss_{lum}$, results in environment maps that are faithful to the original color intensity distribution.
\begin{equation}\label{eq:reg_l1}
\Loss_{L1}(\envmap) = \lambda_{L1}\sum_{i,j} \sum_{k \in \{\text{r,g,b}\}} |\envmap_{i,j}[k]|.
\end{equation}

\minisection{Preserving the Perceptual Features}
To ensure that the synthesized environment maps retain the global perceptual qualities of the original, we utilized the Deep Image Structure and Texture Similarity (DISTS) metric as an additional regularization term $\Loss_{dists} = \lambda_{dists} DISTS(\envmap, \envmap_0)$. The DISTS metric \cite{ding2020iqa} evaluates similarity by focusing on the structural and textural aspects of images, leveraging deep learning to compare these perceptual elements rather than just pixel-wise differences. When used as a regularization term, it guides the optimization towards more natural-looking images.

%-------------------------------------------------------------------------
\section{Experiments}\label{sec:expt}
We run the inverse rendering optimization described at \cref{fig:main_opt} with different SIREN training strategies. All the experiments ran on a single RTX 4090 GPU. The environment maps are taken from PolyHaven \cite{polyhaven}, resized to $256 \times 128$ to speed up the computation time, and cropped to the upper part, which is the only one contributing to the test scenes with the infinite plane. The environment maps are stored in the OpenEXR format, using 32-bit floating-point numbers. The scenes are rendered using the Mitsuba 3\cite{Jakob2020DrJit} differentiable renderer.

\begin{figure}[tb]
  \centering
  \resizebox{\linewidth}{!}{%
    \begin{tabular}{@{}c@{\hskip 0.5em}c@{\hskip 0.5em}c@{\hskip 0.5em}c@{}}
    \vspace{-0.2em}
        \raisebox{0.135\linewidth}{\rotatebox[origin=c]{90}{\scriptsize Initial}} &
        \begin{subfigure}{.35\linewidth} 
            \includegraphics[width=\linewidth]{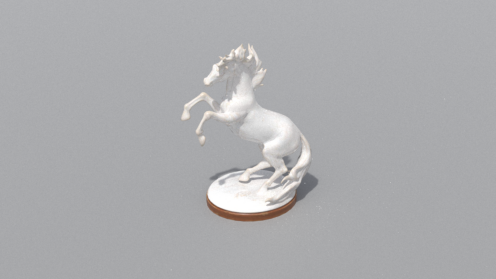}\par
            \includegraphics[width=\linewidth]{opt_results/envmap1/initial_envmap.png}
        \end{subfigure} & 
        \raisebox{0.135\linewidth}{\rotatebox[origin=c]{90}{\scriptsize Ground Truth}} &
        \begin{subfigure}{.35\linewidth} 
            \includegraphics[width=\linewidth]{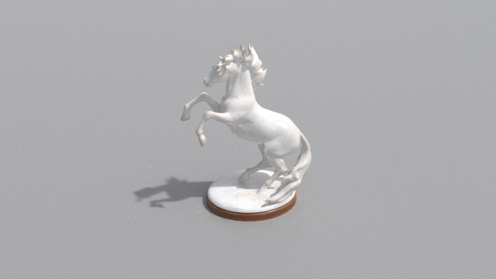}\par
            \includegraphics[width=\linewidth]{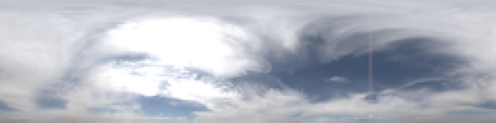}
        \end{subfigure}\\
        \vspace{-0.2em}
        \raisebox{0.135\linewidth}{\rotatebox[origin=c]{90}{\scriptsize Pixels (baseline)}} &
        \begin{subfigure}{.35\linewidth} 
            \includegraphics[width=\linewidth]{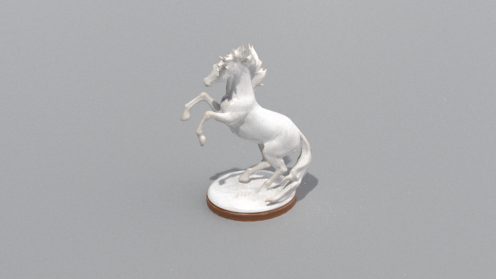}\par
            \includegraphics[width=\linewidth]{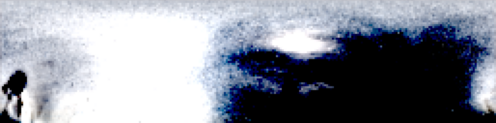}
        \end{subfigure} & 
        \raisebox{0.135\linewidth}{\rotatebox[origin=c]{90}{\scriptsize Log(pixels)}} &
        \begin{subfigure}{.35\linewidth} 
            \includegraphics[width=\linewidth]{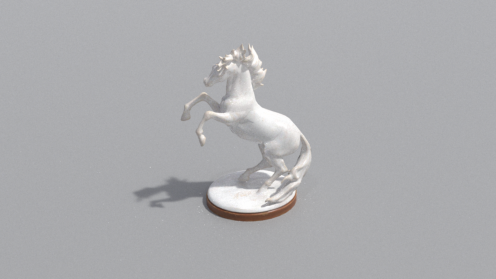}\par
            \includegraphics[width=\linewidth]{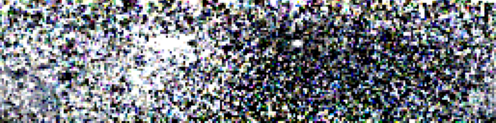}
        \end{subfigure}\\
        \vspace{-0.2em}
        \raisebox{0.135\linewidth}{\rotatebox[origin=c]{90}{\scriptsize SIREN HDR w/o $\Loss_{\text{hdr}}$}} &
        \begin{subfigure}{.35\linewidth} 
            \includegraphics[width=\linewidth]{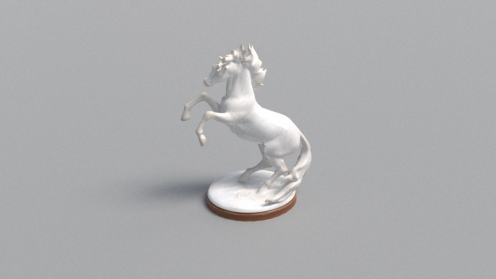}\par
            \includegraphics[width=\linewidth]{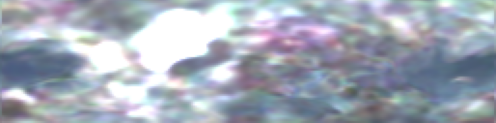}
        \end{subfigure} & 
        \raisebox{0.135\linewidth}{\rotatebox[origin=c]{90}{\scriptsize SIREN HDR}} &
        \begin{subfigure}{.35\linewidth} 
            \includegraphics[width=\linewidth]{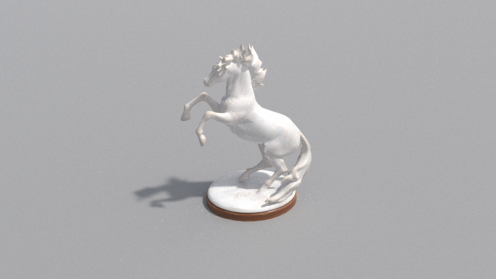}\par
            \includegraphics[width=\linewidth]{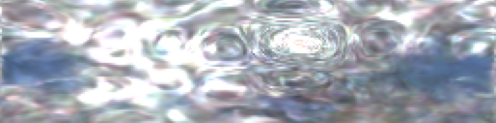}
        \end{subfigure}\\
        \vspace{-0.2em}
        \raisebox{0.135\linewidth}{\rotatebox[origin=c]{90}{\scriptsize R-SIREN HDR}} &
        \begin{subfigure}{.35\linewidth} 
            \includegraphics[width=\linewidth]{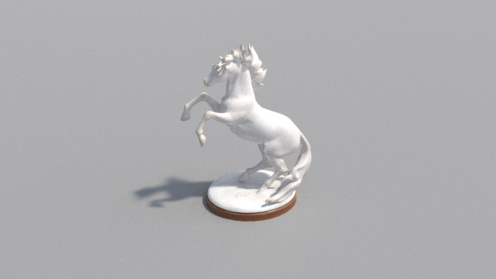}\par
            \includegraphics[width=\linewidth]{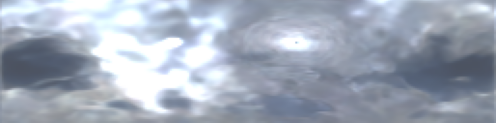}
        \end{subfigure} & 
        \raisebox{0.135\linewidth}{\rotatebox[origin=c]{90}{\scriptsize R-SIREN HDR $\Loss_{\text{reg}}$} \hspace{-0.4em}} &
        \begin{subfigure}{.35\linewidth} 
            \includegraphics[width=\linewidth]{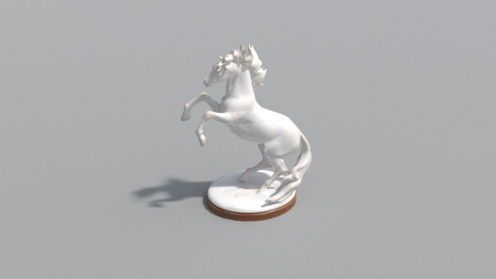}\par
            \includegraphics[width=\linewidth]{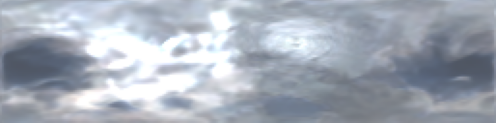}
        \end{subfigure}\\
    \end{tabular}
  }
  \caption{Results over many SIREN training strategies and experiment setups. The initial rendering is obtained with the initial environment map. The goal is to run the optimization to reach the target's lighting conditions. Note that the target environment map is unknown during the optimization.}
\label{fig:reshorse}
  
\end{figure}

\begin{figure}[b]
  \centering
  \resizebox{\linewidth}{!}{%
    \begin{tabular}{@{}c@{\hskip 0.5em}c@{\hskip 0.5em}c@{\hskip 0.5em}c@{}}
        % -- ROW 1 --
        \raisebox{0.118\linewidth}{\rotatebox[origin=c]{90}{\scriptsize Initial Rendering}} &
        \begin{subfigure}{.35\linewidth} 
            \begin{subfigure}{.5\linewidth}
                \begin{tikzpicture}[
                    zoomboxarray,
                    zoomboxarray columns=1,
                    zoomboxarray rows=1,
                    connect zoomboxes,
                    zoombox paths/.append style={line width=1pt}
                ]
                \node [image node] {\adjincludegraphics[width=0.977\linewidth,trim={2.6cm 0.3cm 2.8cm 0.3cm},clip]{opt_results/envmap1/initial_rendering_sphere.png} };
                \zoombox[magnification=3]{0.5,0.62}
            \end{tikzpicture}
        \end{subfigure}\par
        \includegraphics[width=1\linewidth]{opt_results/envmap1/initial_envmap.png}
        \end{subfigure} & 
        \raisebox{0.118\linewidth}{\rotatebox[origin=c]{90}{\scriptsize Ground Truth}} &
        \begin{subfigure}{.35\linewidth} 
            \begin{subfigure}{.5\linewidth}
                \begin{tikzpicture}[
                    zoomboxarray,
                    zoomboxarray columns=1,
                    zoomboxarray rows=1,
                    connect zoomboxes,
                    zoombox paths/.append style={line width=1pt}
                ]
                \node [image node] {\adjincludegraphics[width=0.977\linewidth,trim={2.6cm 0.3cm 2.8cm 0.3cm},clip]{opt_results/envmap1/target_rendering_sphere.png} };
                \zoombox[magnification=3]{0.5,0.62}
            \end{tikzpicture}
        \end{subfigure}\par
        \includegraphics[width=1\linewidth]{opt_results/envmap1/target_envmap.png}
        \end{subfigure}\\
        % -- ROW 2 --
        \raisebox{0.135\linewidth}{\rotatebox[origin=c]{90}{\scriptsize Pixels (baseline)}} &
        \begin{subfigure}{.35\linewidth} 
            \begin{subfigure}{.5\linewidth}
                \begin{tikzpicture}[
                    zoomboxarray,
                    zoomboxarray columns=1,
                    zoomboxarray rows=1,
                    connect zoomboxes,
                    zoombox paths/.append style={line width=1pt}
                ]
                \node [image node] {\adjincludegraphics[width=0.977\linewidth,trim={2.6cm 0.3cm 2.8cm 0.3cm},clip]{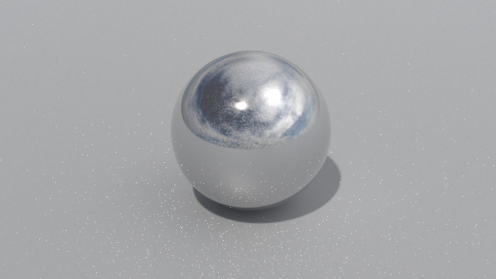} };
                \zoombox[magnification=3]{0.5,0.62}
            \end{tikzpicture}
        \end{subfigure}\par
        \includegraphics[width=1\linewidth]{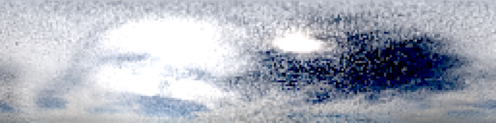}
        \end{subfigure} & 
        \raisebox{0.135\linewidth}{\rotatebox[origin=c]{90}{\scriptsize Log(pixels)}} &
        \begin{subfigure}{.35\linewidth} 
            \begin{subfigure}{.5\linewidth}
                \begin{tikzpicture}[
                    zoomboxarray,
                    zoomboxarray columns=1,
                    zoomboxarray rows=1,
                    connect zoomboxes,
                    zoombox paths/.append style={line width=1pt}
                ]
                \node [image node] {\adjincludegraphics[width=0.977\linewidth,trim={2.6cm 0.3cm 2.8cm 0.3cm},clip]{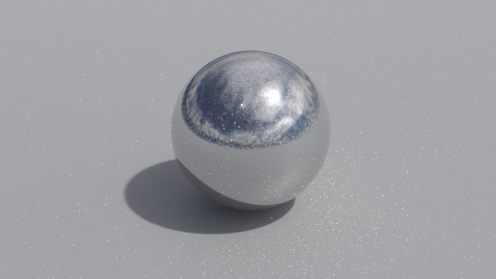} };
                \zoombox[magnification=3]{0.5,0.62}
            \end{tikzpicture}
        \end{subfigure}\par
        \includegraphics[width=1\linewidth]{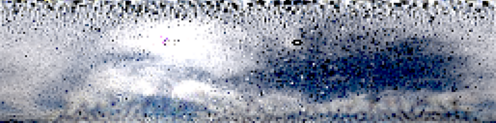}
        \end{subfigure}\\
        % -- ROW 3 --
        \raisebox{0.128\linewidth}{\rotatebox[origin=c]{90}{\scriptsize SIREN HDR}} &
        \begin{subfigure}{.35\linewidth} 
            \begin{subfigure}{.5\linewidth}
                \begin{tikzpicture}[
                    zoomboxarray,
                    zoomboxarray columns=1,
                    zoomboxarray rows=1,
                    connect zoomboxes,
                    zoombox paths/.append style={line width=1pt}
                ]
                \node [image node] {\adjincludegraphics[width=0.977\linewidth,trim={2.6cm 0.3cm 2.8cm 0.3cm},clip]{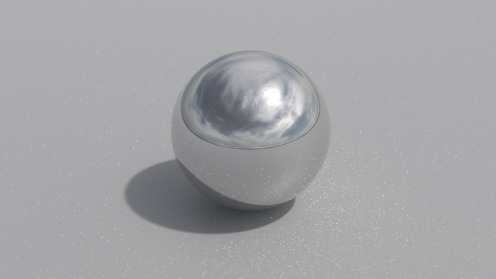} };
                \zoombox[magnification=3]{0.5,0.62}
            \end{tikzpicture}
        \end{subfigure}\par
        \includegraphics[width=1\linewidth]{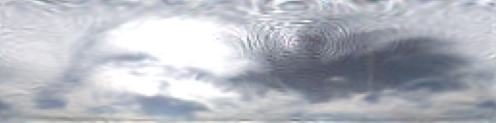}
        \end{subfigure} & 
        \raisebox{0.128\linewidth}{\rotatebox[origin=c]{90}{\scriptsize R-SIREN HDR $\Loss_{reg}$} \hspace{-0.4em}} &
        \begin{subfigure}{.35\linewidth} 
            \begin{subfigure}{.5\linewidth}
                \begin{tikzpicture}[
                    zoomboxarray,
                    zoomboxarray columns=1,
                    zoomboxarray rows=1,
                    connect zoomboxes,
                    zoombox paths/.append style={line width=1pt}
                ]
                \node [image node] {\adjincludegraphics[width=0.977\linewidth,trim={2.6cm 0.3cm 2.8cm 0.3cm},clip]{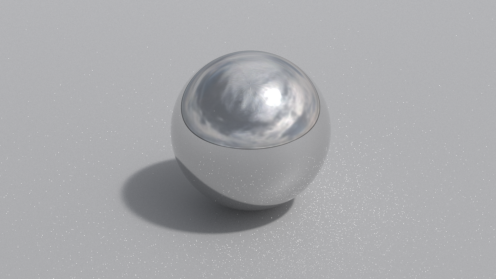} };
                \zoombox[magnification=3]{0.5,0.62}
            \end{tikzpicture}
        \end{subfigure}\par
        \includegraphics[width=1\linewidth]{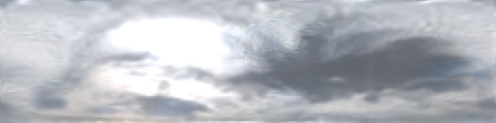}
        \end{subfigure}\\
    \end{tabular}
  }
\caption{Accuracy in the reproduction of the reflections for a full mirror metallic sphere with roughness set to zero. R-SIREN obtains natural-looking environment maps with fewer artifacts.}
\label{fig:spheres}
\end{figure}

\subsection{R-SIREN HDR Implementation Details}
We trained the SIREN models for $1500$ epochs, using the Adam optimizer with $lr=5\times 10^{-5}$, averaging 30 seconds of training time per image. For the Adversarial Weight Perturbation, we employed a proxy Adam optimizer of $1\times 10^{-4}$, applying perturbations of size $\gamma = 0.01$. Both learning rates are scheduled with a Cosine Annealing scheduler with Warm Restarts, where the number of iterations before the first restart is $T_0 = 100$. We do not enforce a lat-long prior at this step, since it is automatically handled by the inverse rendering process.

For the model architecture, compared to stock SIREN, which terminates the neural network with a linear layer, we appended a final Sigmoid activation function as the last layer to strictly enforce the results inside the $(0, 1)$ interval to properly undo the normalization, according to \cref{eq:siren_pred}. The model is then instantiated with $2$ input features, $3$ output features, $256$ hidden features, and $6$ hidden layers. The sinusoidal activation layers have the initial angular frequency set to $\omega_0 = 30$. Our choice of hyperparameters for the loss terms is $\lambda_{\text{log\_mse}} = 0.85$, $\lambda_{\text{log\_ssim}} = 0.25$, and $\lambda_{\text{hdr}} = 1\times 10^{-2}$.

\subsection{Inverse Rendering}
We get the HDR prediction from SIREN with \cref{eq:eps_normexp}. We then use the SSIM loss to enforce the similarity in the shadows, regularized with luminance and L1, according to \cref{eq:reg_lum} and \cref{eq:reg_l1}, and the DISTS metric.
\begin{multline}
\Loss = \lambda_{ssim}\Loss_{SSIM}(I_{new},I_{target}) +\\+ \underbrace{\lambda_{lum} \Loss_{lum}(\envmap_{\text{pred}}, \envmap_0) +\lambda_{dists} \Loss_{dists}(\envmap_{\text{pred}}, \envmap_0) + \lambda_{\text{L1}} \Loss_{L1}(\envmap_{\text{pred}})}_{\Loss_{\text{reg}}}
\end{multline}
With $\lambda_{ssim} \in [1,1\times 10^4]$, $\lambda_{L1} \in [1\times 10^{-4}, 1\times 10^{-8}]$,  $\lambda_{lum} \in [1\times 10^{-4}, 1\times 10^{-8}]$, and $\Loss_{dists} = 0.6$.
Initially, we employ higher $\lambda_{L1}$ and $\lambda_{lum}$ to escape the local minima and reduce the original light intensity. Subsequently, we gradually decrease the regularization weights for L1 and luminance, while simultaneously increasing the emphasis on SSIM, to ensure accurate reconstruction of the shadows. We keep the DISTS hyperparameter weight constant to preserve the original perceptual features at each step. We used the Adam optimizer, with $lr=5\times 10^{-6}$, running for $400$ iterations.

%-------------------------------------------------------------------------
\subsection{Evaluation}
\begin{figure}[!h]
\centering
\captionsetup[subfigure]{width=0.8\textwidth}
\captionsetup[figure]{width=0.8\linewidth}
\centering
\par\smallskip
\begin{subfigure}{.455\columnwidth}
  \includegraphics[width=\linewidth]{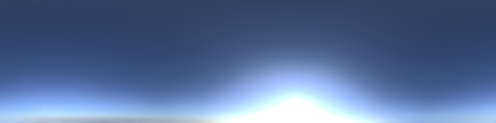}
  \caption*{Initial}
\end{subfigure}
\begin{subfigure}{.455\columnwidth}
  \includegraphics[width=\linewidth]{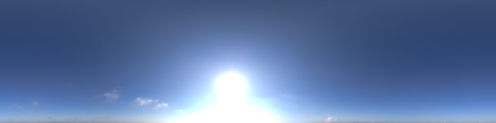}
  \caption*{Ground Truth}  
\end{subfigure}\par\vspace{0.35cm}

\begin{subfigure}{.3\columnwidth}
  \begin{overpic}[width=\linewidth,trim={2.6cm 0.3cm 2.8cm 0.3cm},clip]{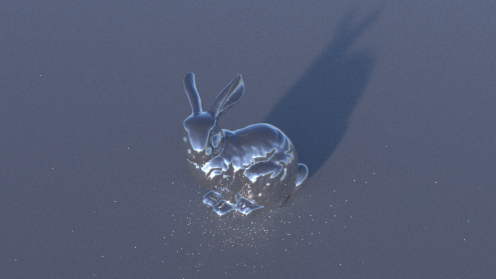}
    \put(-15,30){\rotatebox{90}{Initial}} % First row
  \end{overpic}\par\vspace{0.05cm}
  \begin{overpic}[width=\linewidth,trim={2.6cm 0.3cm 2.8cm 0.3cm},clip]{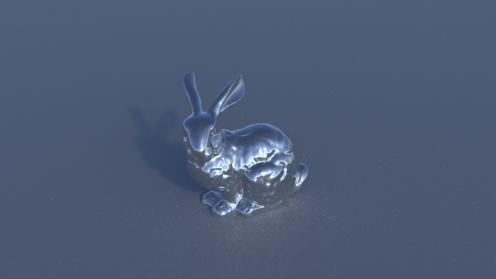}
    \put(-15,30){\rotatebox{90}{Target}} % Second row
  \end{overpic}\par\vspace{0.05cm}
  
  \begin{overpic}[width=\linewidth,trim={2.6cm 0.3cm 2.8cm 0.3cm},clip]{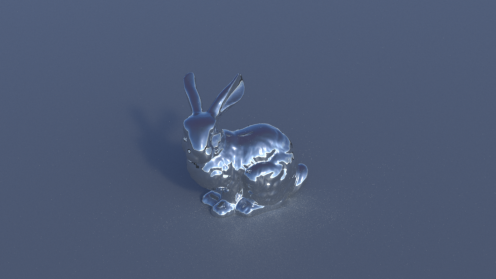}
    % Third row
  \end{overpic}\par\vspace{0.025cm}
  
  \begin{overpic}[width=\linewidth]{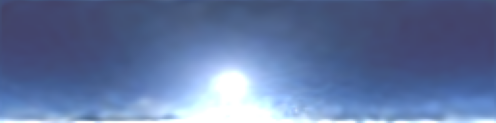}
    \put(-15,40){\rotatebox{90}{Result}} % Third and fourth rows combined
  \end{overpic}
    \caption*{$R=0.0$}
\end{subfigure}
\begin{subfigure}{.3\columnwidth}
  \begin{overpic}[width=\linewidth,trim={2.6cm 0.3cm 2.8cm 0.3cm},clip]{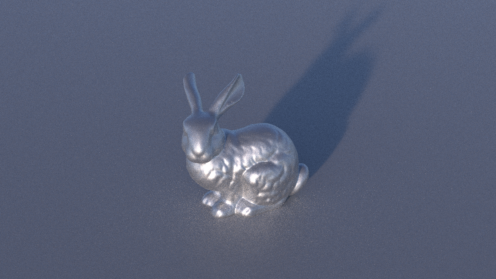}
    \put(-15,30){\rotatebox{90}{}} % First row
  \end{overpic}\par\vspace{0.05cm}
  \begin{overpic}[width=\linewidth,trim={2.6cm 0.3cm 2.8cm 0.3cm},clip]{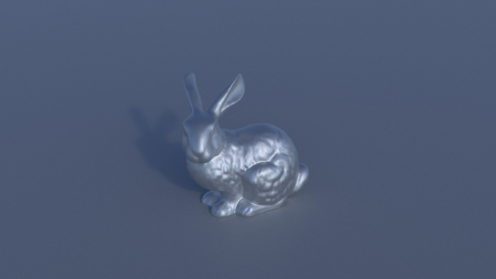}
    \put(-15,30){\rotatebox{90}{}} % Second row
  \end{overpic}\par\vspace{0.05cm}
  
  \begin{overpic}[width=\linewidth,trim={2.6cm 0.3cm 2.8cm 0.3cm},clip]{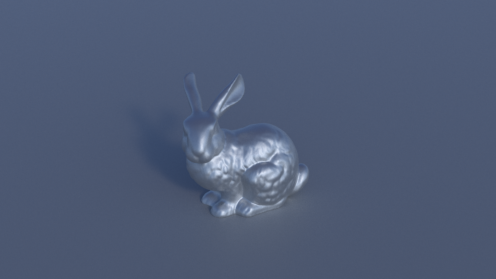}
  % Third row
  \end{overpic}\par\vspace{0.025cm}
  
  \begin{overpic}[width=\linewidth]{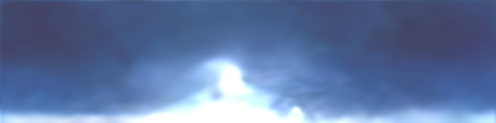}
    \put(-15,40){\rotatebox{90}{}} % Third and fourth rows combined
  \end{overpic}
    \caption*{$R=0.5$}
\end{subfigure}
\begin{subfigure}{.3\columnwidth}
  \begin{overpic}[width=\linewidth,trim={2.6cm 0.3cm 2.8cm 0.3cm},clip]{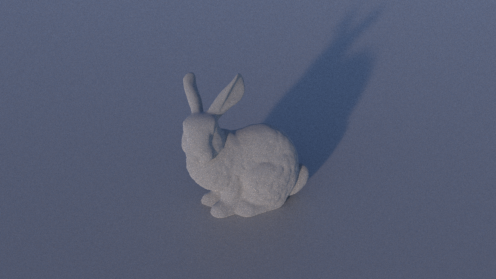}
    \put(-15,30){\rotatebox{90}{}} % First row
  \end{overpic}\par\vspace{0.05cm}
  
  \begin{overpic}[width=\linewidth,trim={2.6cm 0.3cm 2.8cm 0.3cm},clip]{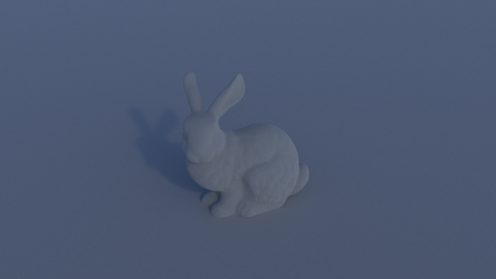}
    \put(-15,30){\rotatebox{90}{}} % Second row
  \end{overpic}\par\vspace{0.05cm}
  
  \begin{overpic}[width=\linewidth,trim={2.6cm 0.3cm 2.8cm 0.3cm},clip]{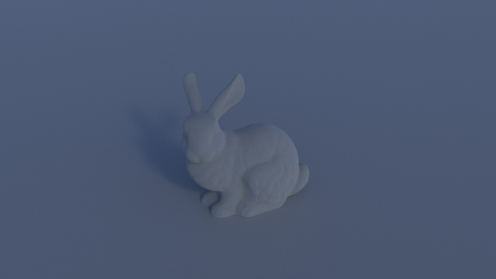}
% Third row
  \end{overpic}\par\vspace{0.025cm}
  
  \begin{overpic}[width=\linewidth]{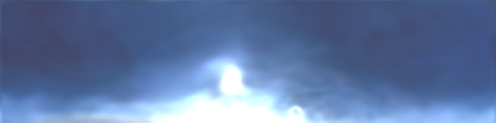}
    \put(-15,40){\rotatebox{90}{}} % Third and fourth rows combined
  \end{overpic}
    \caption*{$R=1.0$}
\end{subfigure}
\caption{Inverse rendering with roughness $R$ set to 0, 0.5, 1, using R-SIREN HDR $\Loss_{reg}$. The ground truth is a different environment map in a different light orientation. Despite the roughness to 1 impairs the inverse rendering process, our method is still able to recover an environment map close to the ground truth.}
\label{fig:bunny}
\end{figure}

\begin{table*}[tb]
    \caption{Evaluation of rendering and environment map against target and ground truth, for the experiment presented in \cref{fig:reshorse}.}
    \label{table:evalhorse}
    
    \centering
    \small
    \begin{tabularx}{\textwidth}{X 
    >{\raggedleft\arraybackslash}p{1.6cm} 
    >{\raggedleft\arraybackslash}p{1.3cm}
    >{\raggedleft\arraybackslash}p{1.3cm} 
    >{\raggedleft\arraybackslash}p{1.3cm} 
    >{\raggedleft\arraybackslash}p{1.6cm} 
    >{\raggedleft\arraybackslash}p{1.3cm} 
    >{\raggedleft\arraybackslash}p{1.3cm} 
    >{\raggedleft\arraybackslash}p{1.3cm}}
        \toprule
        & \multicolumn{4}{c}{\textbf{Envmap Metrics}} & \multicolumn{4}{c}{\textbf{Rendering Metrics}} \\
        \cmidrule(lr){2-5} \cmidrule(lr){6-9}
        \textbf{Experiment} & \textbf{$\uparrow$ PSNR} & \textbf{$\downarrow$ LPIPS} & \textbf{$\downarrow$ MSE} & \textbf{$\uparrow$ SSIM} & \textbf{$\uparrow$ PSNR} & \textbf{$\downarrow$ LPIPS} & \textbf{$\downarrow$ MSE} & \textbf{$\uparrow$ SSIM} \\
        \midrule
Pixels                 &  8.91931 dB &  0.63838  &  0.12825  &  0.31881  &  29.91687 dB &  0.17168  &  0.00102  &  0.94152  \\
Log(Pixels)          &  8.32278 dB &  0.76582  &  0.14714  &  0.02603  &  46.27205 dB &  0.06250  &  \textbf{0.00002}  &  0.99040  \\
SIREN HDR w/o $\Loss_{hdr}$ &  \textbf{14.89818} dB &  0.60814  &  \textbf{0.03237}  &  0.47149  &  38.95401 dB &  0.08466  &  0.00013  &  0.98292  \\
SIREN HDR              &  14.56471 dB &  0.62082  &  0.03496  &  0.34456  &  \textbf{46.37604 dB} &  \textbf{0.05282}  &  \textbf{0.00002}  &  \textbf{0.99316}  \\
R-SIREN HDR       &  14.58689 dB &  0.55724  &  0.03478  &  \textbf{0.55372}  &  39.95935 dB &  0.05812  &  0.00010  &  0.99125  \\
R-SIREN HDR $\Loss_{reg}$   &  14.85838 dB &  \textbf{0.53229}  &  0.03267  &  0.52657  &  41.23402 dB &  0.09429  &  0.00008  &  0.98908  \\
        \bottomrule
    \end{tabularx}
\end{table*}

\begin{table*}[tb]
    \caption{Evaluation of rendering and environment map against target and ground truth, for the experiment presented in \cref{fig:spheres}.}
    \label{table:evalsphere}
    
    \centering
    \small
    
    \begin{tabularx}{\textwidth}{X 
    >{\raggedleft\arraybackslash}p{1.6cm} 
    >{\raggedleft\arraybackslash}p{1.3cm}
    >{\raggedleft\arraybackslash}p{1.3cm} 
    >{\raggedleft\arraybackslash}p{1.3cm} 
    >{\raggedleft\arraybackslash}p{1.6cm} 
    >{\raggedleft\arraybackslash}p{1.3cm} 
    >{\raggedleft\arraybackslash}p{1.3cm} 
    >{\raggedleft\arraybackslash}p{1.3cm}}
        \toprule
        & \multicolumn{4}{c}{\textbf{Envmap Metrics}} & \multicolumn{4}{c}{\textbf{Rendering Metrics}} \\
        \cmidrule(lr){2-5} \cmidrule(lr){6-9}
        \textbf{Experiment} & \textbf{$\uparrow$ PSNR} & \textbf{$\downarrow$ LPIPS} & \textbf{$\downarrow$ MSE} & \textbf{$\uparrow$ SSIM} & \textbf{$\uparrow$ PSNR} & \textbf{$\downarrow$ LPIPS} & \textbf{$\downarrow$ MSE} & \textbf{$\uparrow$ SSIM} \\
        \midrule
Pixels        &  18.35015 dB &  0.48241  &  0.01462  &  0.54777 &  26.01747 dB &  0.25495  &  0.00250  &  0.88970  \\
Log(Pixels) &  14.60996 dB &  0.63089  &  0.03459  &  0.28612 &  30.00922 dB &  \textbf{0.05187}  &  0.00100  &  0.97107  \\
SIREN HDR $\Loss_{reg}$              &  29.19136 dB &  0.32992  &  0.00120  &  0.85988 &  \textbf{40.78696 dB} &  0.09896  &  \textbf{0.00008}  &  \textbf{0.98072}  \\
R-SIREN HDR $\Loss_{reg}$       &  \textbf{29.77839 dB} &  \textbf{0.29141}  &  \textbf{0.00105}  &  \textbf{0.91918} &  35.46278 dB &  0.17063  &  0.00028  &  0.95863  \\
        \bottomrule
    \end{tabularx}
\end{table*}

\minisection{Qualitative} 
To evaluate the representation capabilities of R-SIREN HDR, we started with a physically correct target to avoid biases from the renderer. The first experiment involved reconstructing a new environment map starting from a rendered target in different lighting conditions. In this case, the initial environment map shifted to the left.~\cref{fig:reshorse} shows the results of the preliminary test. 

\newcommand\bestzero[1]{\textcolor{RedViolet}{\textbf{#1}}}
\newcommand\besthalf[1]{\textcolor{PineGreen}{\textbf{#1}}}
\newcommand\bestone[1]{\textcolor{Blue}{\textbf{#1}}}
\begin{table*}[tb]
    \caption{Comprehensive Evaluation of All Puresky Maps from PolyHaven in the Bunny Scene \cite{polyhaven}. The colors indicate the best result for each level of roughness employed. The proposed R-SIREN HDR $\Loss_{reg}$ achieves the best performance among all the metrics for the environment map while still fitting well the target (rendering metrics) across different roughness values.}
    \label{table:evalglobal}
    
    \centering
    \small
    \begin{tabularx}{\textwidth}{X
    >{\centering\arraybackslash}p{1.5cm} 
    >{\raggedleft\arraybackslash}p{1.6cm} 
    >{\raggedleft\arraybackslash}p{1.2cm}
    >{\raggedleft\arraybackslash}p{1.2cm} 
    >{\raggedleft\arraybackslash}p{1.2cm} 
    >{\raggedleft\arraybackslash}p{1.6cm} 
    >{\raggedleft\arraybackslash}p{1.2cm} 
    >{\raggedleft\arraybackslash}p{1.2cm} 
    >{\raggedleft\arraybackslash}p{1.2cm}}
        \toprule
        %--- Header
        \multicolumn{2}{c}{\textbf{Experiment}} & \multicolumn{4}{c}{\textbf{Envmap Metrics}} & \multicolumn{4}{c}{\textbf{Rendering Metrics}} \\
        %--- Sub-Header
        \cmidrule(lr){1-2} \cmidrule(lr){3-6} \cmidrule(lr){7-10}
        \textbf{Name} & \textbf{Roughness} & \textbf{$\uparrow$ PSNR} & \textbf{$\downarrow$ LPIPS} & \textbf{$\downarrow$ MSE} & \textbf{$\uparrow$ SSIM} & \textbf{$\uparrow$ PSNR} & \textbf{$\downarrow$ LPIPS} & \textbf{$\downarrow$ MSE} & \textbf{$\uparrow$ SSIM} \\

    \midrule

\multirow{3}{*}{\shortstack[c]{Pixels (baseline)}}  & \bestzero{0.0} & 11.47963 dB & 0.63746 & \bestzero{0.07877} & 0.33030 & \bestzero{35.00420} & \bestzero{0.18879} & \bestzero{0.00465} & 0.94913  \\ & \besthalf{0.5} & 9.50746 dB & 0.68670 & 0.12130 & 0.23377 & 34.86679 & 0.20606 & \besthalf{0.00507} & 0.95656  \\ & \bestone{1.0} & 9.80608 dB & 0.66478 & 0.11211 & 0.28438 & 36.98002 & 0.20773 & \bestone{0.00337} & 0.95282  \\

\cmidrule(lr){1-10}\multirow{3}{*}{\shortstack[c]{Log(pixels)}}  & \bestzero{0.0} & \bestzero{12.65512 dB} & 0.66607 & 0.09697 & 0.36745 & 31.23284 & 0.32984 & 0.04663 & 0.79454  \\ & \besthalf{0.5} & 9.39572 dB & 0.72085 & 0.16545 & 0.28348 & 24.51719 & 0.54301 & 0.08933 & 0.69031  \\ & \bestone{1.0} & 11.01039 dB & 0.69084 & 0.13392 & \bestone{0.33982} & 29.61368 & 0.45936 & 0.05712 & 0.74746  \\

\cmidrule(lr){1-10}\multirow{3}{*}{\shortstack[c]{SIREN HDR $\Loss_{reg}$}}  & \bestzero{0.0} & 11.52815 dB & 0.48079 & 0.08656 & 0.35959 & 30.14610 & 0.24235 & 0.03281 & 0.93911  \\ & \besthalf{0.5} & 11.18299 dB & 0.51395 & 0.09276 & 0.29486 & 36.25536 & 0.21043 & 0.03165 & 0.95327  \\ & \bestone{1.0} & 10.91892 dB & 0.52043 & 0.09769 & 0.28518 & 38.52983 & 0.21901 & 0.03354 & 0.95579  \\

\cmidrule(lr){1-10}\multirow{3}{*}{\shortstack[c]{R-SIREN HDR $\Loss_{reg}$}}  & \bestzero{0.0} & 11.77427 dB & \bestzero{0.47657} & 0.08324 & \bestzero{0.37889} & 31.73236 & 0.20716 & 0.01330 & \bestzero{0.95850}  \\ & \besthalf{0.5} & \besthalf{11.37157 dB} & \besthalf{0.50795} & \besthalf{0.08931} & \besthalf{0.31936} & \besthalf{38.55692} & \besthalf{0.16639} & 0.01403 & \besthalf{0.97359}  \\ & \bestone{1.0} & \bestone{11.09265 dB} & \bestone{0.51780} & \bestone{0.09469} & 0.30140 & \bestone{40.64959} & \bestone{0.17940} & 0.01541 & \bestone{0.97294}  \\

    \bottomrule
\end{tabularx}
\end{table*}
\begin{figure*}[tb]
  \centering
  \resizebox{\linewidth}{!}{%
    \begin{tabular}{@{}c@{\hskip 0.15em}c@{\hskip 0.15em}c@{\hskip 0.15em}c@{\hskip 0.15em}c@{}}
        Initial Scene & Target & No regularization & $\Loss_{lum}+\Loss_{l1}$ & $\Loss_{lum}+\Loss_{l1}+\Loss_{dists}$ \\
        % -- ROW 1 --
        \begin{subfigure}{0.2\linewidth} 
            \begin{subfigure}{\linewidth}
            \adjincludegraphics[width=1\linewidth,trim={2.6cm 0.3cm 2.8cm 0.3cm},clip]{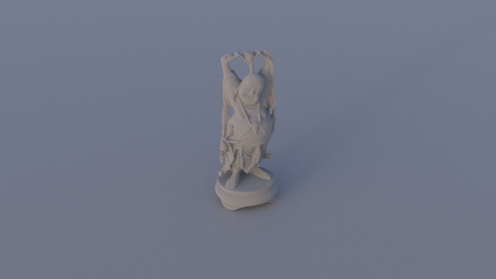}
            \end{subfigure}\par
            \includegraphics[width=\linewidth]{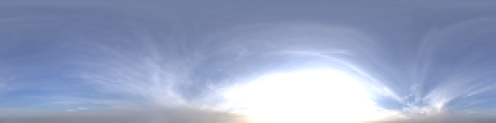}
        \end{subfigure} & 
        % -- ROW 1 --
        \begin{subfigure}{0.2\linewidth} 
            \begin{subfigure}{\linewidth}
            \adjincludegraphics[width=1\linewidth,trim={2.6cm 0.3cm 2.8cm 0.3cm},clip]{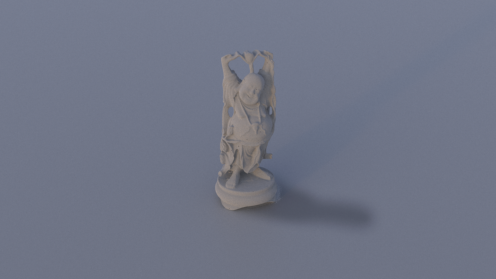}
            \end{subfigure}\par
            \includegraphics[width=\linewidth]{opt_results/placeholder.png}
        \end{subfigure} & 
        % -- ROW 1 --
        \begin{subfigure}{0.2\linewidth} 
            \begin{subfigure}{\linewidth}
            \adjincludegraphics[width=1\linewidth,trim={2.6cm 0.3cm 2.8cm 0.3cm},clip]{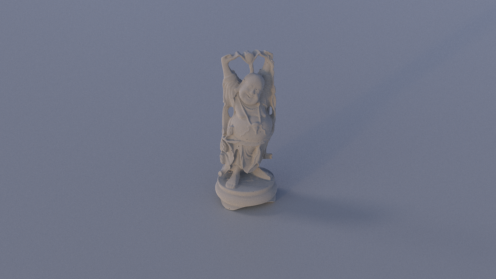}
            \end{subfigure}\par
            \includegraphics[width=\linewidth]{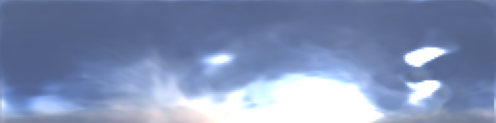}
        \end{subfigure} & 
        % -- ROW 1 --
        \begin{subfigure}{0.2\linewidth} 
            \begin{subfigure}{\linewidth}
            \adjincludegraphics[width=1\linewidth,trim={2.6cm 0.3cm 2.8cm 0.3cm},clip]{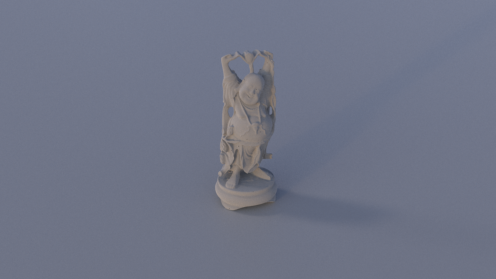}
            \end{subfigure}\par
            \includegraphics[width=\linewidth]{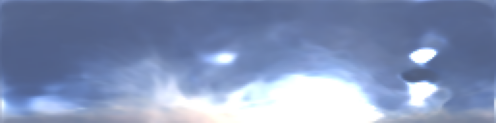}
        \end{subfigure} & 
        % -- ROW 1 --
        \begin{subfigure}{0.2\linewidth} 
            \begin{subfigure}{\linewidth}
            \adjincludegraphics[width=1\linewidth,trim={2.6cm 0.3cm 2.8cm 0.3cm},clip]{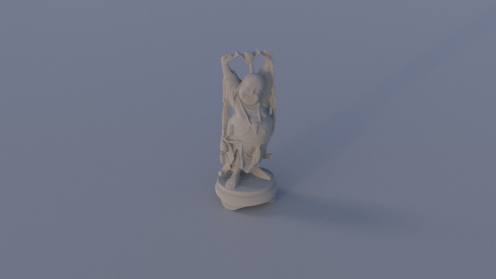} 
            \end{subfigure}\par
            \includegraphics[width=\linewidth]{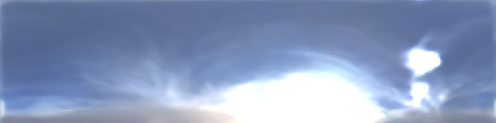} 
        \end{subfigure} \\
        \noalign{\vskip 2mm}  
       
    \end{tabular}
  }
\caption{Effects of the regularization terms over the downstream task. \textit{R-SIREN HDR} already achieves remarkable results without regularization. In addition, employing $\Loss_{lum}+\Loss_{l_1}$ results in sharper shadows due to constraining the brightest pixel in a small region. On the other hand, $\Loss_{lum}+\Loss_{l_1}+\Loss_{dists}$ produces slightly blurrier shadows but preserves more of the perceptual features in the environment map.}
\label{fig:reg_effects_downstream}
\end{figure*}

To assess the reflection quality in our sphere scene, we employed a full mirror as the reflective surface. The experiment is conducted with the SIREN HDR and R-SIREN HDR models to evaluate their quality against the baseline optimization, with Adam directly altering the pixel values. As we can see in \cref{fig:spheres}, both models outperform the baseline optimization in the pixel space, including when the optimizer is altering the values in the logarithmic space. Furthermore, R-SIREN HDR provides finer details in the reconstruction quality.

We further investigated the impact of surface roughness on the inversion process. By varying the roughness values of the reflective surface, we observed that surfaces with minor roughness generally yielded the best reconstructions, as shown in ~\cref{fig:bunny}. This is attributed to the fact that minor roughness facilitates the inverse rendering process. The smoother surface provides more predictable and consistent reflections, making it easier for the inverse rendering process to discover the surrounding environment texture.

\minisection{Quantitative} To provide a more objective assessment of R-SIREN HDR's performance, we employed quantitative metrics such as Peak Signal-to-Noise Ratio (PSNR), Learned Perceptual Image Patch Similarity (LPIPS) \cite{8578166}, Mean Squared Error (MSE), and Structural Similarity Index (SSIM) \cite{1284395}. These metrics measure noise ratio, perceptual similarity, and reconstruction similarity. We measured the quality of both rendering and environment maps against the ground truth for the Mirror Sphere and the Horse experiments, with ~\cref{table:evalhorse} and ~\cref{table:evalsphere}, showing the computed scores.  
The results display a significant gap in the environment map metrics in favor of the SIREN-based approaches. The discrepancy in the results between the environment map and the rendering metrics proves that the unconstrained optimizations lead to overfitting the target scene without focusing on the overall image quality, resulting in final rendering that is very close to the target but degrading the visual features of the environment map. In contrast, LPIPS is a perceptual metric that prioritizes high-level visual features and global similarity, making it less sensitive to local noise and pixel-level discrepancies. Our approach with SIREN is not designed to reconstruct the unknown ground truth by overfitting it, but rather to synthesize a new one while preserving the aesthetic similarity with the original environment map, minimizing noise, and maintaining a continuous structure. This objective is rewarded by the LPIPS, as it shows that R-SIREN HDR $\Loss_{reg}$ has the best capabilities in synthesizing more natural-looking images compared to the pixel-based baselines.

Additionally, Table~\ref{table:evalglobal} presents the evaluation results for all 44 PureSky maps from PolyHaven \cite{polyhaven} on the bunny scene. These results assess the impact of varying roughness on reconstruction quality. When the roughness is set to zero, the baseline approach rapidly overfits the desired sample, guided by the scene's reflection on the object. However, the low LPIPS score indicates that the synthesized environment map lacks the perceptual fidelity of the original, revealing the method's tendency to cheat by using shortcuts in order to fit the target. This is further supported by the low Peak Signal-To-Noise Ratio, demonstrating the baseline's production of noisy images.

Furthermore, removing the reflection information quickly compromises the baseline process, leading to optimization failure. In contrast, our approach achieves superior LPIPS and PSNR scores at lower roughness, resulting in the best performance for generating natural images that accurately capture the structure of the final rendering, as evidenced by the LPIPS and SSIM scores in the rendering metrics.

\minisection{Limitations} While R-SIREN HDR demonstrates promising performance without relying on strong and computationally expensive priors, it shows certain limitations posed by both the single-view inverse rendering and the absence of a latent space. The lack of a strong prior, such as a latent space, makes the model more prone to visual hallucinations, especially when the environment map is not a sky. The model's sensitivity to the training process of SIREN introduces a trade-off between robustness capabilities and sharpness. Additionally, the method's reliance on reflective materials for effective inversion can be challenging in downstream tasks where more reflections require the renderer to establish physical consistency between existing reflections and new sketched lighting conditions. Furthermore, when applying R-SIREN HDR to a downstream task, the user's input specification is crucial. Strokes that do not align with the underlying geometry can lead to physically incorrect and difficult-to-interpret results for the renderer.

%-------------- Downstream task
\begin{figure*}[tb]
  \centering
  \resizebox{\linewidth}{!}{%
    \begin{tabular}{@{}c@{\hskip 0.15em}c@{\hskip 0.15em}c@{\hskip 0.15em}c@{\hskip 0.15em}c@{}}
        Initial Scene & Target & Result & Target & Result \\
        % -- ROW 1 --
        \begin{subfigure}{0.2\linewidth} 
            \begin{subfigure}{\linewidth}
            \adjincludegraphics[width=1\linewidth,trim={2.6cm 0.3cm 2.8cm 0.3cm},clip]{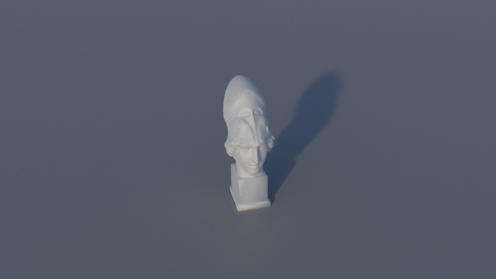}
            \end{subfigure}\par
            \includegraphics[width=\linewidth]{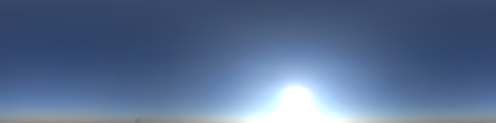}
        \end{subfigure} & 
        % -- ROW 1 --
        \begin{subfigure}{0.2\linewidth} 
            \begin{subfigure}{\linewidth}
            \adjincludegraphics[width=1\linewidth,trim={2.6cm 0.3cm 2.8cm 0.3cm},clip]{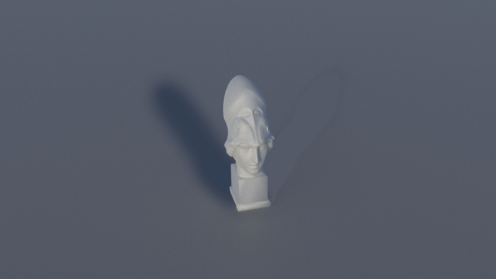}
            \end{subfigure}\par
            \includegraphics[width=\linewidth]{opt_results/placeholder.png}
        \end{subfigure} & 
        % -- ROW 1 --
        \begin{subfigure}{0.2\linewidth} 
            \begin{subfigure}{\linewidth}
            \adjincludegraphics[width=1\linewidth,trim={2.6cm 0.3cm 2.8cm 0.3cm},clip]{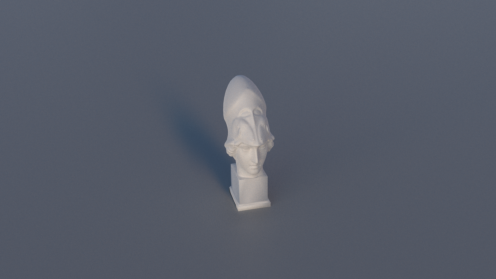}
            \end{subfigure}\par
            \includegraphics[width=\linewidth]{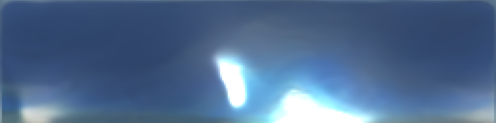}
        \end{subfigure} & 
        % -- ROW 1 --
        \begin{subfigure}{0.2\linewidth} 
            \begin{subfigure}{\linewidth}
            \adjincludegraphics[width=1\linewidth,trim={2.6cm 0.3cm 2.8cm 0.3cm},clip]{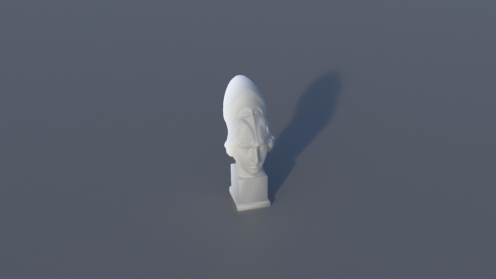}
            \end{subfigure}\par
            \includegraphics[width=\linewidth]{opt_results/placeholder.png}
        \end{subfigure} & 
        % -- ROW 1 --
        \begin{subfigure}{0.2\linewidth} 
            \begin{subfigure}{\linewidth}
            \adjincludegraphics[width=1\linewidth,trim={2.6cm 0.3cm 2.8cm 0.3cm},clip]{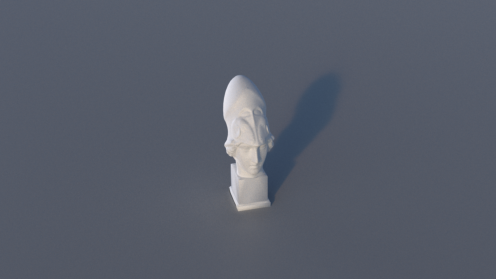} 
            \end{subfigure}\par
            \includegraphics[width=\linewidth]{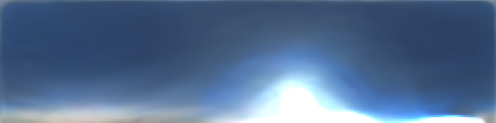} 
        \end{subfigure} \\
        Atenea & & Moving the Shadow & & Adding Reflections\\
        \noalign{\vskip 2mm}  
        % -- ROW 2 --
        \begin{subfigure}{0.2\linewidth} 
            \begin{subfigure}{\linewidth}
            \adjincludegraphics[width=1\linewidth,trim={2.6cm 0.3cm 2.8cm 0.3cm},clip]{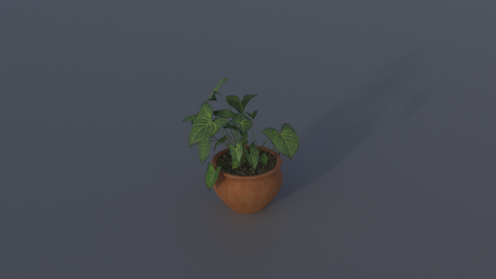}
            \end{subfigure}\par
            \includegraphics[width=\linewidth]{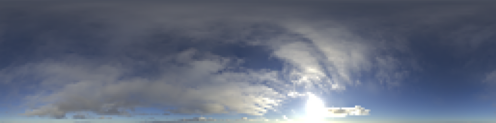}
        \end{subfigure} & 
        % -- ROW 2 --
        \begin{subfigure}{0.2\linewidth} 
            \begin{subfigure}{\linewidth}
            \adjincludegraphics[width=1\linewidth,trim={2.6cm 0.3cm 2.8cm 0.3cm},clip]{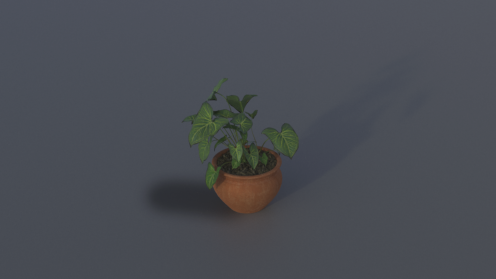}
            \end{subfigure}\par
            \includegraphics[width=\linewidth]{opt_results/placeholder.png}
        \end{subfigure} & 
        % -- ROW 2 --
        \begin{subfigure}{0.2\linewidth} 
            \begin{subfigure}{\linewidth}
            \adjincludegraphics[width=1\linewidth,trim={2.6cm 0.3cm 2.8cm 0.3cm},clip]{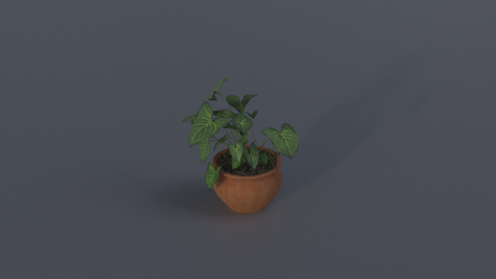}
            \end{subfigure}\par
            \includegraphics[width=\linewidth]{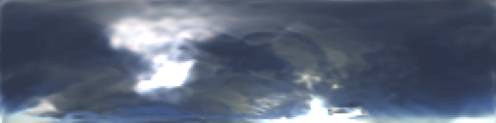}
        \end{subfigure} & 
        % -- ROW 2 --
        \begin{subfigure}{0.2\linewidth} 
            \begin{subfigure}{\linewidth}
            \adjincludegraphics[width=1\linewidth,trim={2.6cm 0.3cm 2.8cm 0.3cm},clip]{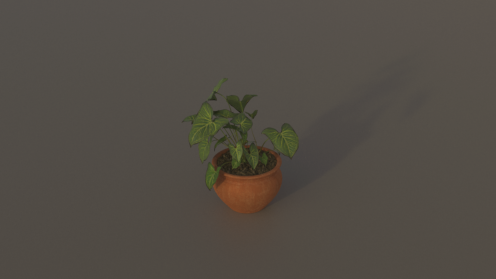}
            \end{subfigure}\par
            \includegraphics[width=\linewidth]{opt_results/placeholder.png}
        \end{subfigure} & 
        % -- ROW 2 --
        \begin{subfigure}{0.2\linewidth} 
            \begin{subfigure}{\linewidth}
            \adjincludegraphics[width=1\linewidth,trim={2.6cm 0.3cm 2.8cm 0.3cm},clip]{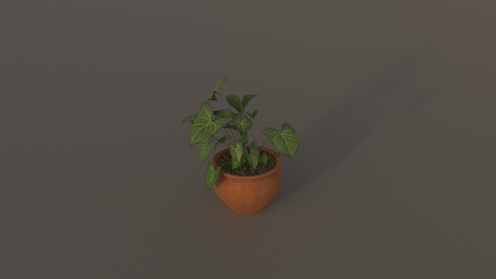} 
            \end{subfigure}\par
            \includegraphics[width=\linewidth]{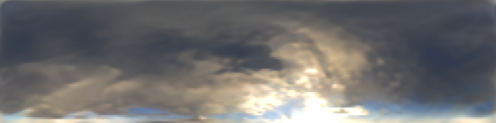} 
        \end{subfigure} \\
        Plant & & Adding Shadows & & Color Temperature\\
        \noalign{\vskip 2mm}  
        % -- ROW 3 --
        \begin{subfigure}{0.2\linewidth} 
            \begin{subfigure}{\linewidth}
            \adjincludegraphics[width=1\linewidth,trim={2.6cm 0.3cm 2.8cm 0.3cm},clip]{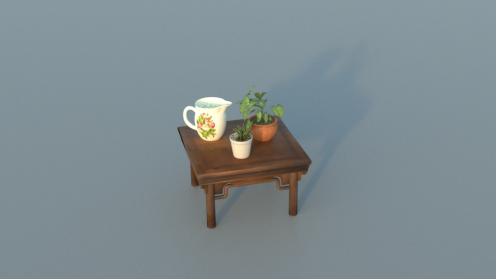}
            \end{subfigure}\par
            \includegraphics[width=\linewidth]{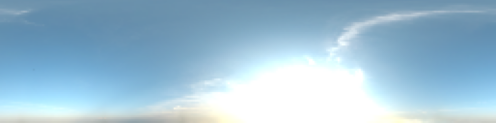}
        \end{subfigure} & 
        % -- ROW 3 --
        \begin{subfigure}{0.2\linewidth} 
            \begin{subfigure}{\linewidth}
            \adjincludegraphics[width=1\linewidth,trim={2.6cm 0.3cm 2.8cm 0.3cm},clip]{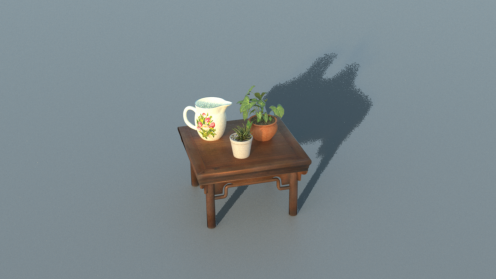}
            \end{subfigure}\par
            \includegraphics[width=\linewidth]{opt_results/placeholder.png}
        \end{subfigure} & 
        % -- ROW 3 --
        \begin{subfigure}{0.2\linewidth} 
            \begin{subfigure}{\linewidth}
            \adjincludegraphics[width=1\linewidth,trim={2.6cm 0.3cm 2.8cm 0.3cm},clip]{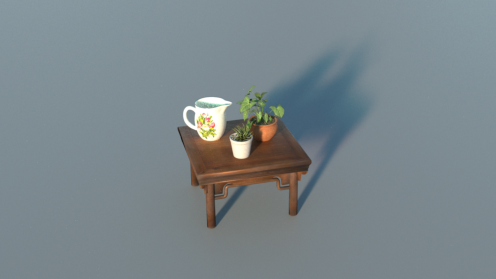}
            \end{subfigure}\par
            \includegraphics[width=\linewidth]{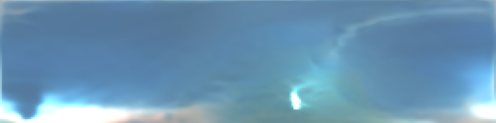}
        \end{subfigure} & 
        % -- ROW 3 --
        \begin{subfigure}{0.2\linewidth} 
            \begin{subfigure}{\linewidth}
            \adjincludegraphics[width=1\linewidth,trim={2.6cm 0.3cm 2.8cm 0.3cm},clip]{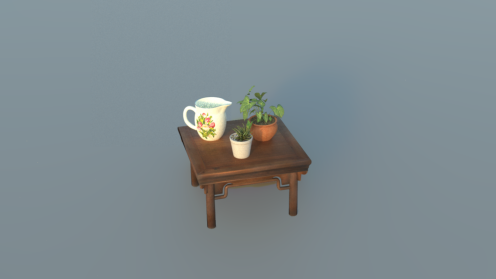}
            \end{subfigure}\par
            \includegraphics[width=\linewidth]{opt_results/placeholder.png}
        \end{subfigure} & 
        % -- ROW 3 --
        \begin{subfigure}{0.2\linewidth} 
            \begin{subfigure}{\linewidth}
            \adjincludegraphics[width=1\linewidth,trim={2.6cm 0.3cm 2.8cm 0.3cm},clip]{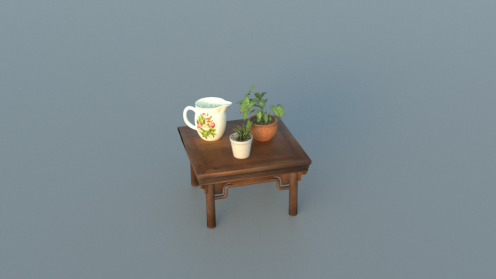} 
            \end{subfigure}\par
            \includegraphics[width=\linewidth]{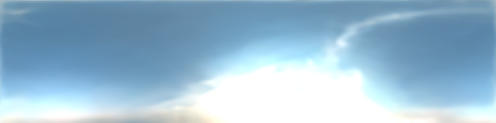} 
        \end{subfigure} \\
        Multi-Objects Textured & & Enhancing the Shadow & & Blurring the Shadow\\
        \noalign{\vskip 2mm}  
        
    \end{tabular}
  }
\caption{Downstream task. Synthesized environment maps and renderings with various appearance edits: $\diamond$ \textit{Moving the Shadow}: the existing shadow is painted over with the background color, and then a new one is sketched in the desired location. $\diamond$ \textit{Adding Reflections}: the new reflection coming from the light is painted over the 3D model. $\diamond$ \textit{Adding Shadows}: the new shadow is sketched without removing the original ones. $\diamond$ \textit{Color temperature}: the color temperature of the image is altered to a warmer appearance. $\diamond$ \textit{Enhancing the Shadow}: the shadow is coarsely selected and then increased in contrast.  $\diamond$ \textit{Blurring the Shadow}: a Gaussian blur filter is applied within the desired region.}
\label{fig:downstreamtask}
\end{figure*}
%--------------- end

\subsection{Downstream Task: Appearance Editing}
Moving the focus to downstream applications, appearance editing is the task of synthesizing a new environment map based on some user's input over the original rendering. We used a photo editing program to sketch target images over the initial renderings,
capturing our desired lighting conditions. Compared to the previous inverse rendering tasks, we are not computing the loss over a physically correct image because older features, such as reflections over the objects, are still present, creating conflicts that will be blended by the inversion process. The renderer will have to find the best compromise to a physically correct sample that covers all the light conditions expressed in the target. We reduced the luminance regularization weights to give the optimization more freedom, allowing new lights without completely removing older ones. The effects of the regularization terms over the downstream task are illustrated in \cref{fig:reg_effects_downstream}.
We tested various appearance editing methods: shadow translation, light addition, shadow addition, color temperature adjustment, and shadow enhancement/blurring. For shadow translation, we removed the existing shadow by painting over it with the background color and then drew a new shadow in the desired location. To add a new shadow, we applied a dark brush stroke to the desired area. For light addition, we introduced a lighter brush stroke to simulate the desired light reflection on the model. To adjust the color temperature, we applied a color temperature filter. Finally, we enhanced or blurred the shadows by increasing the contrast and applying Gaussian blur, respectively.~\cref{fig:downstreamtask} illustrates the synthesized environment maps and the corresponding final renderings.

%-------------------------------------------------------------------------
\section{Conclusions and Future Work}
In this work, we proposed a method to modify existing environment maps using a lazy-learning optimization technique. This approach leverages the output of a differentiable renderer as a target within the loss function. To preserve the overall brightness characteristics, we incorporated a luminance regularization. Additionally, we used L1 regularization to restrict the HDR light features to small, meaningful sectors, imitating the natural behavior of light distribution in environment maps. Lastly, we applied the DISTS-based regularizer to keep the original perceptual features. We then demonstrated how to transition the image representation from RGB to neural implicit functions, leading to significant improvements in the consistency of the newly generated images. To achieve this, we trained a neural network to represent HDR images and ensured its robustness to small changes in the weights, facilitating optimization guided by the inverse rendering loss. Enforcing robustness during the training of the neural implicit function helped maintain the stability of the overall perceptual features and reduce noise. Finally, we applied the proposed R-SIREN representation to the downstream task of editing the appearance of environment maps by specifying a coarse target crafted with photo editing software. In this way, we were able to obtain novel environment maps such that the final rendering resembles said target, significantly improving the perceptual fidelity and reducing the overall noise. We did so without relying on expensive pre-trained generative models such as Generative Adversarial Networks (GANs) or Denoising Diffusion Probabilistic Models (DDPMs).

In our future research, we aim to explore adversarial implicit functions for 3D model manipulation via inverse rendering, potentially extending to other representations like SDFs. Furthermore, we plan to refine the way we enforce the "local" latent space of neural implicit functions by incorporating diverse samples during perturbations. This includes augmentations of the same sample or introducing new references to bring their representation closer inside the weights space of the trained sample, enabling the appearance editing process to properly blend the different content and perceptual features. By doing so, we expect to enhance the model's ability to generate more realistic and diverse outputs.

\section*{Acknowledgments}
This work was supported by projects PNRR MUR PE0000013-FAIR under the MUR National Recovery and Resilience Plan funded by the European Union - NextGenerationEU and PRIN 2022 project 20227YET9B “AdVVent” CUP
code B53D23012830006.
%-------------------------------------------------------------------------
% bibtex
\bibliographystyle{eg-alpha-doi} 
\bibliography{egbibsample}       

% biblatex with biber
% \printbibliography                

%-------------------------------------------------------------------------
\end{document}